\documentclass[10pt,journal,compsoc]{IEEEtran}
\ifCLASSOPTIONcompsoc
  \usepackage[nocompress]{cite}
\else
\fi
\ifCLASSINFOpdf
\else
\fi
\usepackage{hyperref}
\usepackage{tikz}
\usepackage{times}
\usepackage{epsfig}
\usepackage{graphicx}

\usepackage[numbers,sort&compress]{natbib}
\usepackage{mathrsfs}
\usepackage{amsmath}
\usepackage{amssymb}
\usepackage{colortbl}

\usepackage[ruled,vlined,lines numbered]{algorithm2e}

\usepackage{subfigure}

\usepackage{cite} 
\newcommand{\rmnum}[1]{\romannumeral #1}

\usepackage{epsf}

\usepackage{stackrel}
\usepackage{centernot}

\usepackage{tikz}
\usetikzlibrary{shapes,arrows}
\usetikzlibrary{fit}					
\usetikzlibrary{backgrounds}	

\usepackage{multirow} 
\usepackage{booktabs}

\usepackage{pgfplots}

\usepackage{pgfplots}

\usepackage{datatool}

\newfont{\mycrnotice}{ptmr8t at 7pt}
\newfont{\myconfname}{ptmri8t at 7pt}
\begin{document}
%
\title{A Survey of Multi-View Representation Learning}

\author{Yingming Li, Ming Yang, Zhongfei (Mark) Zhang,~\IEEEmembership{Senior~Member,~IEEE}
\IEEEcompsocitemizethanks{\IEEEcompsocthanksitem Y. Li, M. Yang, Z. Zhang are with College of Information Science \& Electronic Engineering, Zhejiang University, China.\protect\\
E-mail: {\{yingming, cauchym, zhongfei\}@zju.edu.cn}
}
\thanks{}}

%
%

\markboth{Journal of \LaTeX\ Class Files,~Vol.~14, No.~8, August~2015}%
{Shell \MakeLowercase{\textit{et al.}}: Bare Demo of IEEEtran.cls for Computer Society Journals}
%



\IEEEtitleabstractindextext{%
\begin{abstract}
Recently, multi-view representation learning has become a rapidly growing direction in machine learning and data mining areas. This paper introduces two categories for multi-view representation learning: multi-view representation alignment and multi-view representation fusion. Consequently, we first review the representative methods and theories of multi-view representation learning based on the perspective of alignment, such as correlation-based alignment. Representative examples are canonical correlation analysis (CCA) and its several extensions. Then from the perspective of representation fusion we investigate the advancement of multi-view representation learning that ranges from generative methods including multi-modal topic learning, multi-view sparse coding, and multi-view latent space Markov networks, to neural network-based methods including multi-modal autoencoders, multi-view convolutional neural networks, and multi-modal recurrent neural networks. Further, we also investigate several important applications of multi-view representation learning. Overall, this survey aims to provide an insightful overview of theoretical foundation and state-of-the-art developments in the field of multi-view representation learning and to help researchers find the most appropriate tools for particular applications.
\end{abstract}

}

\maketitle

\IEEEdisplaynontitleabstractindextext

%
\IEEEpeerreviewmaketitle

\section{Introduction}
\begin{figure}
\begin{center}
   \includegraphics[scale=0.25]{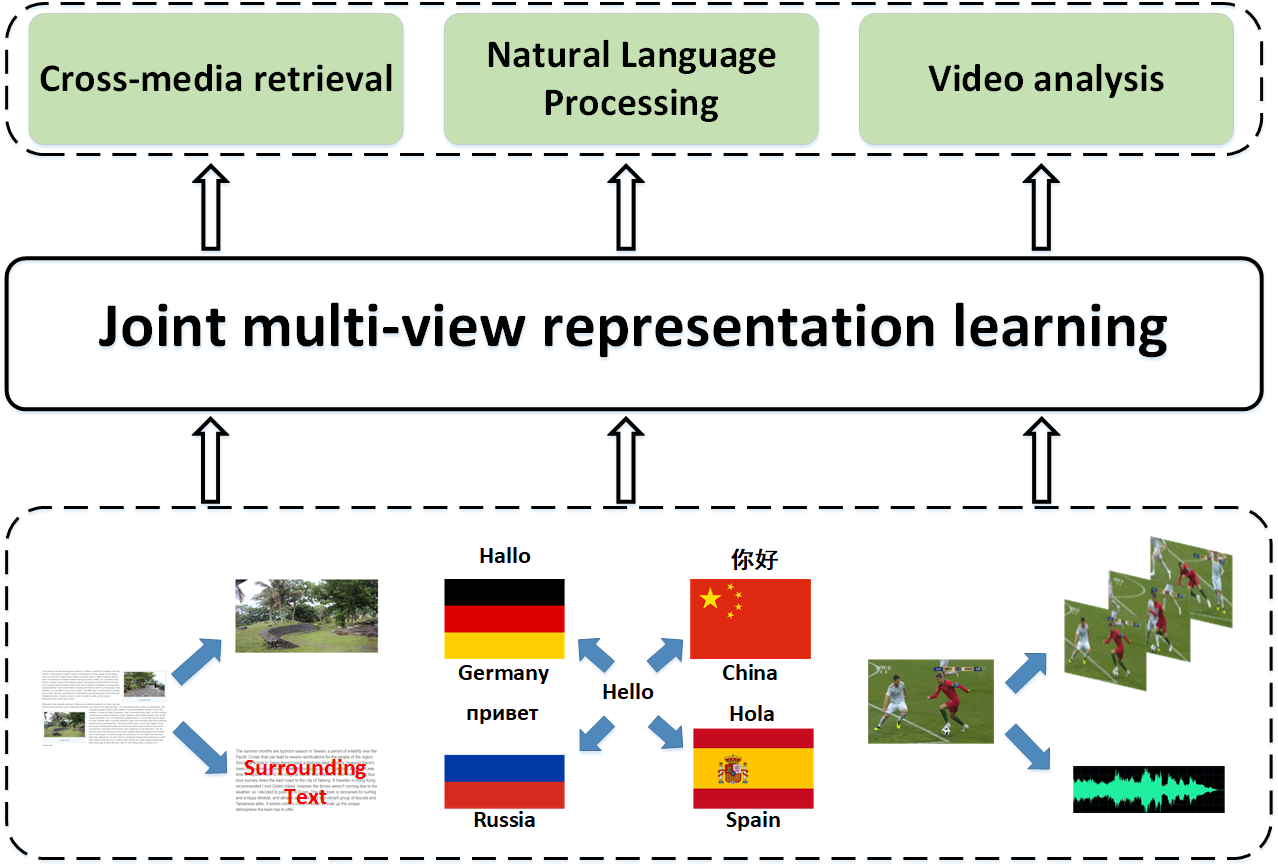}
\end{center}
\vspace*{-0.8em}
\caption{Multi-view data and several related applications based on joint multi-view representation learning.}
\label{motivations}
\vspace*{-0.8em}
\end{figure}

Multi-view representation learning is concerned with the problem of learning representations (or features) of the multi-view data that facilitate extracting readily useful information when developing prediction models. This learning mechanism has attracted much attention since multi-view data have become increasingly available in real-world applications (Figure \ref{motivations}) where examples are described by multi-modal measurements of an underlying signal, such as image+text, audio+video, audio+articulation, and text in different languages, or synthetic views of the unimodal measurements, such as word+context words, different time stamps of a time sequence, and web text+text of inbound hyperlinks. Generally, data from different views usually contain complementary information and multi-view representation learning exploits this point to learn more comprehensive representations than those of single-view learning methods. Since the performance of machine learning methods is heavily dependent on the expressive power of data representation, multi-view representation learning has become a very promising topic with wide applicability.

Canonical correlation analysis (CCA) \cite{Hotelling1936} and its kernel extensions \cite{BachJ02,Hardoon2004,sun2013survey} are representative techniques in early studies of multi-view representation learning. A variety of theories and approaches are later introduced to investigate their theoretical properties, explain their success, and extend them to improve the generalization performance in particular tasks. While CCA and its kernel versions show their abilities of effectively modeling the relationship between two or more sets of variables, they have limitations on capturing high level associations between multi-view data. Inspired by the success of deep neural networks \cite{vincent2008extracting,salakhutdinov2009deep,bengio2013representation}, deep CCAs \cite{AndrewABL13} have been proposed to solve this problem, with a common strategy to learn a joint representation that is coupled between multiple views at a higher level after learning several layers of view-specific features in the lower layers.
\begin{figure*}
\begin{center}
   \includegraphics[scale=0.34]{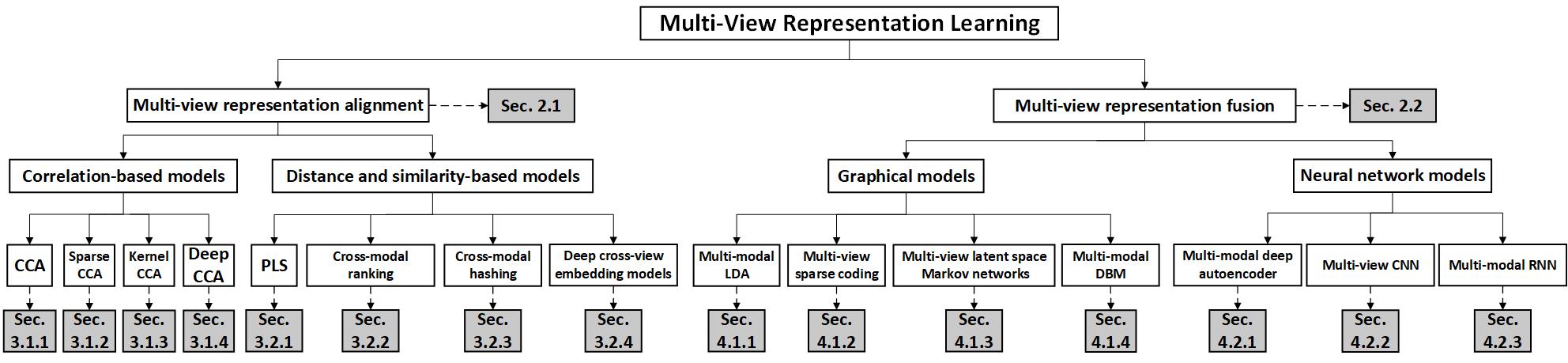}
\end{center}
   \caption{The basic organization of this survey. The left part shows the architecture of multi-view representation learning methods based on the alignment perspective, and the right part displays the structure of multi-view embedding models based on the fusion perspective. }
\label{organization}
\vspace*{-0.8em}
\end{figure*}
However, how to learn a good association between multi-view data still remains an open problem. In 2016, a workshop on multi-view representation learning is held in conjunction with the 33rd international conference on machine learning to help promote a better understanding of various approaches and the challenges in specific applications. So far, there have been increasing research activities in this direction and a large number of multi-view representation learning algorithms have been presented based on the fundamental theories of CCAs and progress of deep neural networks. For example, the advancement of multi-view representation learning ranges from the traditional methods including multi-modal topic learning \cite{CohnH00,Barnard2003,BleiJ03}, multi-view sparse coding \cite{JiaSD10,CaoJMPCN13,liu2014multiview}, and multi-view latent space Markov networks \cite{XingYH05,ChenZX10}, to deep architecture-based methods including multi-modal deep Boltzmann machines \cite{SrivastavaS12}, multi-modal deep autoencoders \cite{ngiam2011multimodal,feng2014cross,WangALB15DCCAE}, and multi-modal recurrent neural networks \cite{KarpathyF14,mao2014deep,donahue2015long}. 

Based on the extensive literature investigation and analysis, we propose two major categories for multi-view representation learning: (\rmnum{1}) multi-view representation alignment, which aims to capture the relationships among multiple different views through feature alignment; (\rmnum{2}) multi-view representation fusion, which seeks to fuse the separate features learned from multiple different views into a single compact representation. Both strategies seek to exploit the complementary knowledge contained in multiple views to comprehensively represent the data.

Consequently, we review the literature of multi-view representation learning based the above taxonomy. Figure \ref{organization} illustrates the organization of this survey. In particular, multi-view representation alignment is surveyed based on different ways of alignment: distance-based, similarity-based, and correlation-based alignment. The underlying idea is that data of each view are processed by a mapping function and then the learned separate representations are regularized with certain constraints to form a multi-view aligned space. On the other hand, multi-view representation fusion is reviewed from probabilistic graphical and neural network perspectives. In fact, the fundamental difference between the two paradigms is whether the layered architecture of a learning model is to be interpreted as a probabilistic graphical model or as a computation network.

The goal of this survey is to review the theoretical foundation and key advances in the area of multi-view representation learning and to provide a global picture of this active direction. We expect this survey to help researchers find the most appropriate approaches for their particular applications and deliver perspectives of what can be done in the future to promote the development of multi-view representation learning.

\subsection{Main Differences from Other Related Surveys}
Recently, several related surveys \cite{Bleiholder2009,AtreyHEK10,sun2013survey,xu2013survey,LahatAJ15} of multi-view learning have been introduced to investigate the theories and applications of the existing multi-view learning algorithms. Among them, the closest efforts to this article are \cite{sun2013survey} and \cite{xu2013survey}. Both of them focus on multi-view learning techniques using the traditional feature learning methods, and give a comprehensive overview of multi-view learning.

The main differences between these two surveys and this survey are concluded as follows. First, this survey focuses on the multi-view representation learning, while the other two surveys concern all the aspects of multi-view learning. Second, this survey provides a more detailed analysis of various multi-view representation learning models from the traditional literature to deep frameworks. In comparison, the other two surveys mainly investigate the traditional multi-view embedding methods and ignore the recent developments of deep neural network-based methods. Third, \cite{xu2013survey} classifies the multi-view learning algorithms into three different settings: co-training style, multiple kernel learning, and subspace learning; the survey of \cite{sun2013survey} provides a comprehensive review of CCA, effectiveness of co-training, and generalization error analysis for co-training and other multi-view learning approaches. In contrast to these two surveys, this survey is formulated as learning multi-view embeddings from alignment and fusion perspectives that helps understand the basic ideas of joint multi-view representation learning. In particular, co-training \cite{blum1998combining} focuses on the multi-view decision level fusion. It trains separate learners on each view and forces the decision of learners to be similar on the same validation examples. Since this survey mainly concentrates on feature level multi-view learning, the co-training related algorithms are not investigated.

\begin{figure*}[htbp]
\centering 
\subfigure[Muti-view representation alignment]
{
	\begin{minipage}{8cm}
	\centering       
	\includegraphics[scale=0.36]{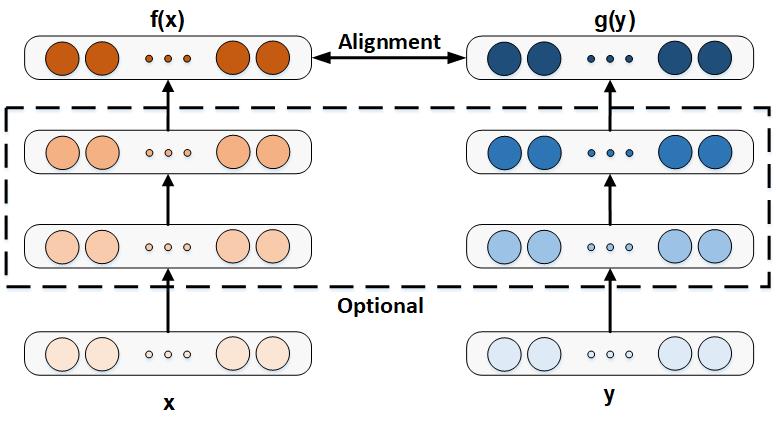} 
	\end{minipage}
	\label{rr1}
}	
\subfigure[Muti-view representation fusion]
{
	\begin{minipage}{8cm}
	\centering     
	\includegraphics[scale=0.36]{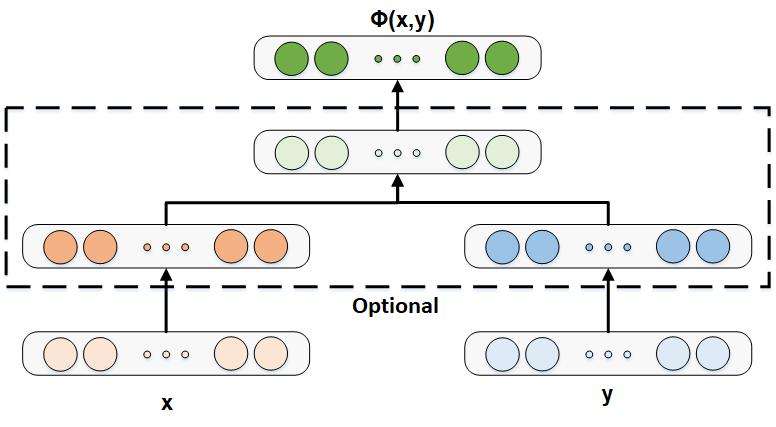}
	\end{minipage}
	\label{rr2}
}
\caption{Multi-view representation learning schemes. The multi-view representation alignment scheme is shown in (a) where the representations from different views are enforced the alignment through certain metrics, such as similarity and distance measurement. The multi-view representation fusion scheme (b) aims to integrate the multi-view inputs into a single and compact representation.} 
\label{taxonomy}
\end{figure*}
\subsection{Challenges for Multi-View Representation Learning}
Many problems have made multi-view representation learning very challenging, including but not limited to (\rmnum{1}) low-quality input data (e.g., noisy and missing values); (\rmnum{2}) inappropriate objectives for multi-view embedding modeling; (\rmnum{3}) scalable processing requirements; (\rmnum{4}) the presence of view disagreement. These challenges may degrade the performance of multi-view representation learning.

A great number of multi-view embedding methods have been proposed to cope with these challenges. These methods usually focus on different issues in multi-view representation learning and thus have different characteristics. Generally, they aim to handle the following questions.      
\begin{itemize}
\item What are the appropriate objectives for learning good multi-view representation?
\item Which types of deep learning architectures are suitable for multi-view representation learning?
\item How should the embedding function be modeled in multi-view representation learning with structured inputs/outputs? 
\item What are the theoretical connections among the different multi-view representation learning paradigms?
\end{itemize}
The response to these questions depends heavily on the particular context of the multi-view learning task and the available multi-view information. Consequently, it is necessary to go into the underlying theoretical principles of the existing literature and discuss in detail the representative multi-view embedding models of each principle. Based on this consideration, we provide a survey to help researchers be aware of several common principles of multi-view representation learning and choose the most suitable models for their particular tasks.

\section{A Taxonomy on Multi-View Representation Learning}
Multi-view representation learning is a scenario of learning representation by relating information of multiple views of the data to boost the learning performance. However, if the learning scenario is not able to coincide with the statistical properties of multi-view data, the obtained representation may even reduce the learning performance. Taking speech recognition with audio and visual data for example, it is difficult to relate raw pixels to audio waveforms, while the two views' data have correlations between mid-level representations, such as phonemes and visemes. 

In this survey, we focus on multi-view joint learning of mid-level representations, where one embedding is introduced for modeling a particular view and then all the embeddings are jointly optimized to leverage the abundant information from multiple views. Through fully investigating the characteristics of the existing successful multi-view representation learning techniques, we propose to divide the current multi-view representation learning methods into two major categories: {\em multi-view representation alignment} and {\em multi-view representation fusion}. An illustration of the two scenarios is shown in Figure \ref{taxonomy}.

\subsection{Multi-View Representation Alignment}
Suppose that we have the two-view given datasets $X$ and $Y$, multi-view representation alignment is expressed as follows: 
\begin{align}
f(x;W_f)\leftrightarrow g(y;W_g)
\end{align}
where each view has a corresponding embedding function ($f$ or $g$) that transforms the original space into a multi-view aligned space with certain constraints and $\leftrightarrow$ denotes the alignment operator.

\noindent\textbf{Distance-based alignment.} A natural {\em distance-based alignment} between the $i$-th pair representation of $x_{i}$ and $y_i$ can be formulated as follows:
\begin{align}
\min_{\theta}||f(x_{i};W_f)-g(y_i;W_g)||_2^2
\end{align}

By extending this alignment constraint, various multi-view representation learning methods have been proposed in decades. For example, Cross-modal Factor Analysis (CFA) is introduced by Li et al. \cite{Li2003MCP} as a simple example of multi-view embedding learning based on the alignment principle. For a given pair $(x_{i},y_{i})$, it aims at finding the orthogonal transformation matrices $W_{x}$ and $W_{y}$ that minimize the following expression:
\begin{align}
||x_{i}^{\top}W_{x}-y_{i}^{\top}W_{y}||^{2}_{2}+r_x(W_x)+r_y(W_y)
\end{align}
where $r_x(\cdot)$ and $r_y(\cdot)$ are regularization terms.

Besides, the idea of distance-based alignment is also applied in multi-view deep representation learning. Correspondence autoencoder \cite{feng2014cross} imposes a distance-based constraint to selected code layers to build correspondence between two views' representations and its loss function on any pair of inputs is defined as follows:
\begin{align}
L=\lambda_{x}\|x_i-\hat{x}_i\|_2^2+&\lambda_{y}\|y_i-\hat{y}_i\|_2^2 \notag\\
&+||f^{c}(x_{i};W_f)-g^{c}(y_i;W_g)||_{2}^{2}
\end{align}
where $f^{c}(x_{i};W_f)$ and $g^{c}(y_i;W_g)$ denote the specific corresponding code layers.

\noindent\textbf{Similarity-based alignment} has also become a popular way to learn aligned spaces. For example, Frome et al. \cite{frome2013devise} introduce a deep visual-semantic embedding model where it encourages a higher dot-product similarity between the visual embedding output and the representation of the correct label than between the visual output and other randomly selected text concepts,
\begin{align}
\sum_{j\neq l}\max\left(0,m-S(t_{l},v_{img})+S(t_{j},v_{img})\right)
\end{align}
where $v_{img}$ is a deep embedding vector for the given image, $t_{l}$ is the learned embedding vector for the provided text label, $t_{j}$ are the embeddings of the other text terms, and $S(\cdot)$ measures the similarity between the two vectors.

Further, Karpathy and Li \cite{KarpathyF14} develop a deep cross-modal alignment model which associates the segments of sentences and the region of an image that they describe through a multi-modal embedding space and a similarity-based structured objective.

\noindent\textbf{Correlation-based alignment} is another typical case of multi-view representation alignment and aims to maximize the correlations of variables among multiple different views through CCA. Given a pair of datasets $X=[x_{1},\ldots,x_{n}]$ and $Y=[y_{1},\ldots,y_{n}]$, Hotelling \cite{Hotelling1936} proposes CCA to find linear projections $w_{x}$ and $w_{y}$, which make the corresponding examples in the two datasets maximally correlated in the projected space,
\begin{align}
\rho=\max_{{w_{x},w_{y}}}\text{corr}\left({w_{x}^{\top}X,w_{y}^{\top}Y}\right)
\end{align}
where $\text{corr}(\cdot)$ denotes the sample correlation function between $w_{x}^{\top}X$ and $w_{y}^{\top}Y$. Thus by maximizing the correlations between the projections of the examples, the basis vectors can be computed for the two sets of variables and applied to two-view data to obtain the required embedding.

Further, the correlation learning can be naturally applied to multi-view neural network learning to learn deep and abstract multi-view representations. For example, deep CCA is proposed by Andrew et al. \cite{AndrewABL13} to obtain deep nonlinear mappings between two views $\{X,Y\}$ which are maximally correlated.

\subsection{Multi-View Representation Fusion}
Suppose that we have the two-view given dataset $X$ and $Y$, multi-view representation fusion is expressed as follows: 
\begin{align}
h=\phi(x,y)
\end{align}
where data from multiple views are integrated into a single representation $h$ which exploits the complementary knowledge contained in multiple views to comprehensively represent the data.

\noindent\textbf{Graphical model-based fusion.} From the generative modeling perspective, the problem of multi-view feature learning can be interpreted as an attempt to learn a compact set of latent random variables that represent a distribution over the observed multi-view data. Under this interpretation, $p(x,y,z)$ can be expressed as a probabilistic model over the joint space of the shared latent variables $z$, and the observed two-view data ${x,y}$. Representation values are determined by the posterior probability $p(z|x,y)$. Representative examples are multi-modal topic learning \cite{BleiJ03}, multi-view sparse coding \cite{JiaSD10}, multi-view latent space Markov networks \cite{XingYH05,ChenZX10}, and multi-modal deep Boltzmann machines \cite{SrivastavaS12}.

Further, taking probabilistic collective matrix factorization (PCMF) \cite{SinghG08,LongZY07,li2010image,GunasekarYYC15} for example, it learns shared multi-view representations over the joint space of multi-view data to fully exploit the complementary information. In particular, a simple form of PCMF considers two views' data matrices $\{X\in\mathbb{R}^{n\times d_{X}},Y\in\mathbb{R}^{n\times d_{Y}}\}$ of the same row dimensionality, and simultaneously factorizes them based on the following probabilistic form
\begin{align}
&p\left(X|\sigma^{2}_{X}\right)=\prod^{n}_{i=1}\mathcal{N}\left(X_{i}|U_{i}V_{X}^{\top},\sigma^{2}_{X}\mathbf{I}\right)\notag\\
&p\left(Y|\sigma^{2}_{Y}\right)=\prod^{n}_{i=1}\mathcal{N}\left(Y_{i}|U_{i}V_{Y}^{\top},\sigma^{2}_{Y}\mathbf{I}\right)
\end{align} 
where $\mathcal{N}(x,|\mu,\sigma^2)$ indicates the Gaussian distribution with mean $\mu$ and variance $\sigma^2$ and the two views' data share the same lower-dimensional factor matrix $U\in\mathbb{R}^{n\times k}$ which can be considered as the common representation. $V_{X}\in\mathbb{R}^{d_{X}\times k}$ and $V_{Y}\in\mathbb{R}^{d_{Y}\times k}$ are the corresponding loading matrices for the two views and are usually placed zero-mean spherical Gaussian priors.

\noindent\textbf{Neural Network-based fusion.} Recently, neural network-based models have been widely used for data representation learning, such as visual and textual data. In particular, deep networks trained on the large datasets have shown their superiority for various tasks such as object recognition \cite{krizhevsky2012}, text classification \cite{kim2014convolutional}, and speech recognition \cite{abdel2014convolutional}. These deep architectures can be adapted to specific domains including multi-view measurements of an underlying signal, such as video analysis and cross-media retrieval. From the neural network modeling perspective, multi-view representation learning first learns the respective mid-level features for each view and then integrates them into a single and compact representation. Representative examples are multi-modal autoencoder \cite{ngiam2011multimodal}, multi-view convolutional neural network \cite{feichtenhofer2016convolutional}, and multi-modal recurrent neural network \cite{KarpathyF14}.

Taking multi-view convolutional neural network for example, it fuses the network at the convolution layer to learn multi-view correspondence feature maps instead of fusing at the softmax layer. Suppose that $\mathbf{x}^{a}$ and $\mathbf{x}^{b}$ are two learned feature maps for views $a$ and $b$ through tied convolution, where weights are shared across the two views. Simple ways of the two-view convolutional feature fusion are as follows, 
\begin{itemize}
\item Sum fusion: $h^{\text{sum}}=\mathbf{x}^{a}+\mathbf{x}^{b}$. 
\item Max fusion: $h^{\text{max}}=\max\{\mathbf{x}^{a},\mathbf{x}^{b}\}$.
\item Concatenation fusion: $h^{\text{cat}}=[\mathbf{x}^{a}, \mathbf{x}^{b}]$.
\end{itemize}
Further, the above fusion strategies are also widely applied to other neural network-based representation fusion methods. For example, Kiela and Bottou \cite{KielaB14} obtain multi-modal concept representations by concatenating a skip-gram linguistic representation vector with a visual concept representation learned with a deep convolutional neural network. Max fusion is usually applied in the setting with synthetic views of the unimodal measurements, such as different time stamps of a time sequence. In video-based person re-identification, McLaughlin et al. \cite{mclaughlin2016recurrent} combine the visual features from all time-stamps using temporal max pooling to obtain an comprehensive appearance feature for the complete sequence. Moreover, Karpathy and Li \cite{KarpathyF14} introduce a multi-modal recurrent neural network to generate image descriptions. This approach learns common multi-modal embeddings for language and visual data and then exploits their complementary information to predict a variable-sized text given an image.

\begin{figure*}
\begin{center}
   \includegraphics[scale=0.35]{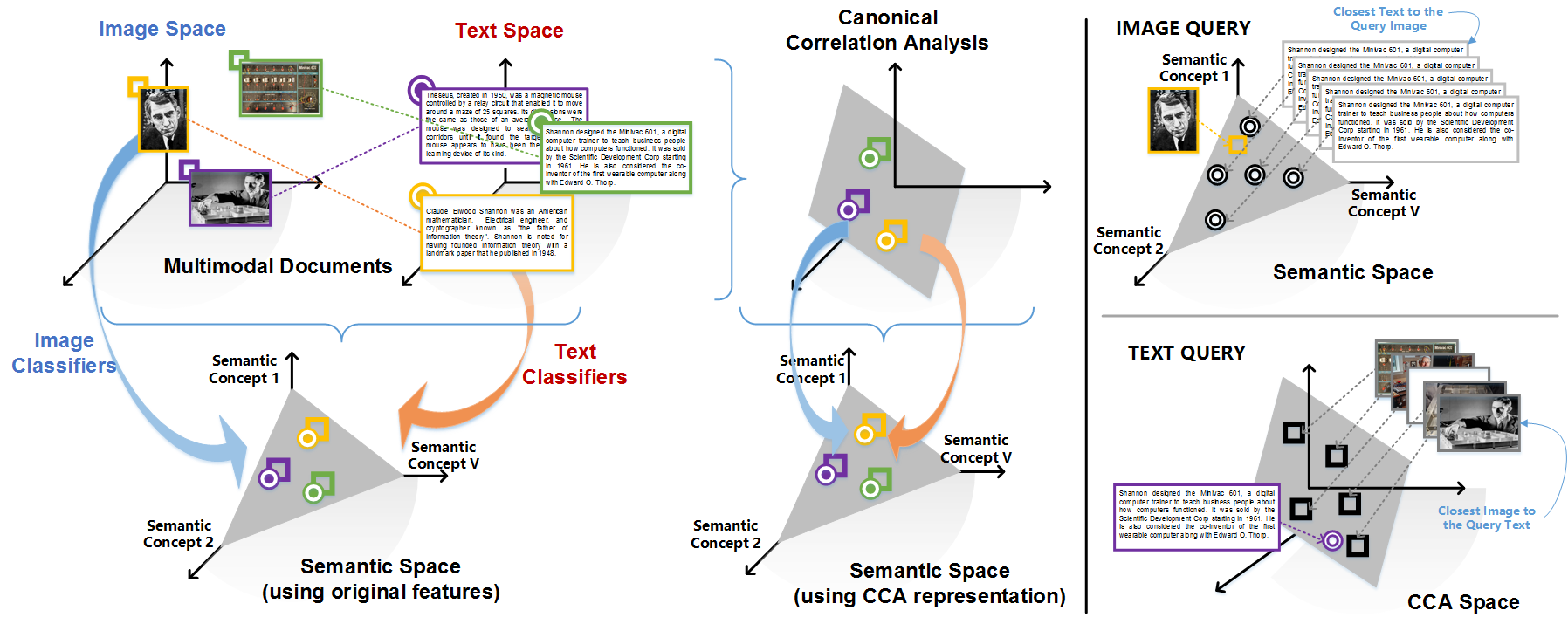}
\end{center}
\caption{An illustrative application example of CCA in cross-modal retrieval (adapted from \cite{Rasiwasia2010}). Left: Embedding of the text and image from their source spaces to a CCA space, Semantic Space and a Semantic space learned using CCA representation. Right: examples of cross-modal retrieval where both text and images are mapped to a common space. At the top is shown an example of retrieving text in response to an image query with a common semantic space. At the bottom is shown an example of retrieving images in response to a text query with a common subspace using CCA.}
\label{cross_CCA1}
\end{figure*}

\section{Multi-View Representation alignment}
Multi-view representation alignment methods seek to perform alignment between the representations learned from multiple different views. Representative examples can be investigated from two aspects: 1) correlation-based alignment; 2) distance and similarity-based alignment. In this section we first review the correlation-based alignment techniques: canonical correlation analysis (CCA) and its extensions that range from the traditional modeling to the non-linear deep embedding. Then typical examples of distance and similarity-based alignment are provided to further show the advantageous of multi-view representation learning.
 
\subsection{Correlation-based Alignment}
In this section we will review the multi-view representation learning techniques from the perspective of correlation-based multi-view alignment: canonical correlation analysis (CCA), sparse CCA, kernel CCA, and deep CCA.
\subsubsection{Canonical Correlation Analysis}
Canonical Correlation Analysis \cite{Hotelling1936} has become increasingly popular for its capability of effectively modeling the relationship between two or more sets of variables. From the perspective of multi-view representation learning, CCA computes a shared embedding of both or more sets of variables through maximizing the correlations among the variables among these sets. More specifically, CCA has been widely used in multi-view learning tasks to generate low-dimensional representations \cite{KidronSE05,Sun2008,Rasiwasia2010}. Improved generalization performance has been witnessed in areas including dimensionality reduction \cite{Foster08multi-viewdimensionality,SunCY10,AvronBTZ13}, clustering \cite{FernBF05,BlaschkoL08,ChaudhuriKLS09}, regression \cite{KakadeF07,McWilliamsBB13}, word embeddings \cite{DhillonFU11,DhillonRFU12,GongKIL14}, and discriminant learning \cite{KimKC07,zhang2011multi,SuFGT12}. For example, Figure \ref{cross_CCA1} shows a fascinating cross-modality application of CCA in cross-media retrieval. 

Given a pair of datasets $X=[x_{1},\ldots,x_{n}]$ and $Y=[y_{1},\ldots,y_{n}]$, CCA tends to find linear projections $w_{x}$ and $w_{y}$, which make the corresponding examples in the two datasets maximally correlated in the projected space. The correlation coefficient between the two datasets in the projected space is given by
\begin{align}
\rho=\text{corr}\left({w_{x}^{\top}X,w_{y}^{\top}Y}\right)=\frac{w_{x}^{\top}C_{xy}w_{y}}{\sqrt{\left(w_{x}^{\top}C_{xx}w_{x}\right)\left(w_{y}^{\top}C_{yy}w_{y}\right)}}
\label{CCA}
\end{align}
where the covariance matrix $C_{xy}$ is defined as
\begin{align}
C_{xy}=\frac{1}{n}\sum_{i=1}^{n}\left(x_{i}-\mu_{x}\right)\left(y_{i}-\mu_{y}\right)^{\top}
\end{align}
where $\mu_{x}=\frac{1}{n}\sum_{i=1}^{n}x_{i}$ and $\mu_{y}=\frac{1}{n}\sum_{i=1}^{n}y_{i}$ are the means of the two views, respectively. The definition of $C_{xx}$ and $C_{yy}$ can be obtained similarly.

Since the correlation $\rho$ is invariant to the scaling of $w_{x}$ and $w_{y}$, CCA can be posed equivalently as a constrained optimization problem.
\begin{align}
&\max_{w_{x},w_{y}} w^{T}_{x}C_{xy}w_{y}\notag\\
\text{s.t.}~~~~~~&w_{x}^{T}C_{xx}w_{x}=1,~~~~w_{y}^{T}C_{yy}w_{y}=1
\label{CCA1}
\end{align}

By formulating the Lagrangian dual of Eq.(\ref{CCA1}), it can be shown that the solution to Eq.(\ref{CCA1}) is equivalent to solving a pair of generalized eigenvalue problems \cite{Hardoon2004},
\begin{align}
C_{xy}C_{yy}^{-1}C_{yx}w_{x}=\lambda^{2}C_{xx}w_{x}\notag\\
C_{yx}C_{xx}^{-1}C_{xy}w_{y}=\lambda^{2}C_{yy}w_{y}
\end{align}

Besides the above definition of CCA, there are also other different ways to define the canonical correlations of a pair of matrices, and all these ways are shown to be equivalent \cite{Golub1992}. In particular, Kettenring \cite{kettenring1971canonical} shows that CCA is equivalent to a constrained least-square optimization problem. Further, Golub and Zha \cite{Golub1992} also provide a classical algorithm for computing CCA by first QR decomposition of the data matrices which whitens the data and then an SVD of the whitened covariance matrix. However, with typically huge data matrices this procedure becomes extremely slow. Avron et al. \cite{AvronBTZ13,AvronBTZ14} propose a fast algorithm for CCA with a pair of tall-and-thin matrices using subsampled randomized Walsh-Hadamard transform \cite{abs-1011-1595}, which only subsamples a small proportion of the training data points to approximate the matrix product. Further, Lu and Foster \cite{LuF14} consider sparse design matrices and introduce an efficient iterative regression algorithm for large scale CCA.

While CCA has the capability of conducting multi-view feature learning and has been widely applied in different fields, it still has some limitations in different applications. For example, it ignores the nonlinearities of multi-view data. Consequently, many algorithms based on CCA have been proposed to extend the original CCA in real-world applications. In the following sections, we review its several widely-used extensions including sparse CCA, kernel CCA, and Deep CCA.

\subsubsection{Sparse CCA}
Recent years have witnessed a growing interest in learning sparse representations of data. Correspondingly, the problem of sparse CCA also has received much attention in the multi-view representation learning. The quest for sparsity can be motivated from several aspects. The first is the ability to account for the predicted results. The big picture usually relies on a small number of crucial variables, with details to be allowed for variation. The second motivation for sparsity is regularization and stability. Reasonable regularization plays an important role in eliminating the influence of noisy data and reducing the sensitivity of CCA to a small number of observations. Further, sparse CCA can be formulated as a subset selection scheme which reduces the dimensionality of the vectors and makes possible a stable solution.
   
The problem of sparse CCA can be considered as finding a pair of linear combinations of $w_x$ and $w_y$ with prescribed cardinality which maximizes the correlation. In particular, sparse CCA can be defined as the solution to
\begin{align}
&\rho=\max_{{w}_{x}, {w}_{y}}\frac{{w}_{x}^{\top}C_{xy}{w}_{y}}{\sqrt{{w}_{x}C_{xx}{w}_{x}{w}_{y}C_{y}{w}_{y}}} \notag \\
\text{s.t.}~~~~~~&||{w}_{x}||_{0}\leq s_{x}, ~~~~~||{w}_{y}||_{0}\leq s_{y}.
\end{align}

Most of the approaches to sparse CCA are based on the well known LASSO trick \cite{tibshirani96regression} which is a shrinkage and selection method for linear regression. By formulating CCA as two constrained simultaneous regression problems, Hardoon and Shawe-Taylor \cite{Hardoon2007} propose to approximate the non-convex constraints with $\infty$-norm. This is achieved by fixing each index of the optimized vector to $1$ in turn and constraining the $1$-norm of the remaining coefficients. Similarly, Waaijenborg et al. \cite{Waaijenborg2008} propose to use the elastic net type regression.

In addition, Sun et al. \cite{Sun2008} introduce a sparse CCA by formulating CCA as a least squares problem in multi-label classification and directly computing it with the Least Angle Regression algorithm (LARS) \cite{Efron04leastangle}. Further, this least squares formulation facilitates the incorporation of the unlabeled data into the CCA framework to capture the local geometry of the data. For example, graph laplacian \cite{Belkin2006} can be used in this framework to tackle with the unlabeled data. 

In fact, the development of sparse CCA is intimately related to the advance of sparse PCA \cite{dAspremontGJL07}. The classical solutions to generalized eigenvalue problem with sparse PCA can be easily extended to that of sparse CCA \cite{Turnbull_NIPS_MBC07,Wiesel2008}. Torres et al. \cite{Turnbull_NIPS_MBC07} derive a sparse CCA algorithm by extending an approach for solving sparse eigenvalue problems using D.C. programming. Based on the sparse PCA algorithm in \cite{dAspremontBG07}, Wiesel et al. \cite{Wiesel2008} propose a backward greedy approach to sparse CCA by bounding the correlation at each stage. Witten et al. \cite{Witten2009} propose to apply a penalized matrix decomposition to the covariance matrix $C_{xy}$, which results in a method for penalized sparse CCA. Consequently, structured sparse CCA has been proposed by extending the penalized CCA with structured sparsity inducing penalty \cite{ChenLC12}.

\subsubsection{Kernel CCA}
Canonical Correlation Analysis is a linear multi-view representation learning algorithm, but for many scenarios of real-world multi-view data revealing nonlinearities, it is impossible for a linear embedding to capture all the properties of the multi-view data \cite{xu2013survey}. Since kernerlization is a principled trick for introducing non-linearity into linear methods, kernel CCA (KCCA) \cite{LaiF00,Akaho01akernel} provides an alternative solution. As a non-linear extension of CCA, KCCA has been successfully applied in many situations, including independent component analysis \cite{BachJ02}, cross-media information retrieval \cite{Hardoon2004,Socher010,HwangG12}, computational biology \cite{YamanishiVNK03,Hardoon2004,BlaschkoSBLG11}, multi-view clustering \cite{BlaschkoL08,Trivedi2010}, acoustic feature learning \cite{AroraL12,AroraL13}, and statistics \cite{BachJ02,GrettonHSBS05}.

The key idea of KCCA lies in embedding the data into a higher dimensional feature space $\phi_{x}:\mathcal{X}\rightarrow\mathcal{H}$, where $\mathcal{H}_{x}$ is the reproducing kernel Hilbert space (RKHS) associated with the real numbers, $k_{x}:\mathcal{X}\times \mathcal{X}\rightarrow \mathbb{R}$ and $k_{x}(x_{i},x_{j})=<\phi_{x}(x_{i}),\phi_{x}(x_{j})>$. $k_{y},\mathcal{H}_{y}$, and $\phi_{y}$ can be defined analogously.

By adapting the representer theorem \cite{learningkernels} to the case of multi-view data to state that the following minimization problem,
\begin{align}
\min_{f_{1},\ldots,f_{k}}&L((x_{1},y_{1},f_{x}(x_{1}),f_{y}(y_{1})),\ldots,(x_{n},y_{n},f_{x}(x_{n}),f_{y}(y_{n})))\notag\\
&+\Omega_{x}(||f||_{K}^{2},||f||_{K}^{2})
\end{align}
where $L$ is an arbitrary loss function and $\Omega$ is a strictly monotonically increasing function, admits representation of the form
\begin{align}
f_{x}(x)=\sum_{i}\alpha_{i}k_{x}(x_{i},x),~~~ f_{y}(y)=\sum_{i}\beta_{i}k_{y}(y_{i},y)
\end{align} 

Correspondingly, we replace vectors $w_{x}$ and $w_{y}$ in our previous CCA formulation Eq.(\ref{CCA}) with $f_{x}=\sum_{i}\alpha_{i}\phi_{x}(x_{i})$ and $f_{y}=\sum_{i}\beta_{i}\phi_{y}(y_{i})$, respectively, and replace the covariance matrices accordingly. The KCCA objective can be written as follows:
\begin{align}
\rho=\frac{f_{x}^{\top}\hat{C}_{xy}f_{y}}{\sqrt{f_{x}^{\top}\hat{C}_{xx}^{\top}f_{x}f_{y}^{\top}\hat{C}_{yy}f_{y}}}
\label{kkcca}
\end{align}
In particular, the kernel covariance matrix $\hat{C}_{xy}$ is defined as
\begin{align}
\hat{C}_{xy}=\frac{1}{n}\sum_{i=1}^{n}\left(\phi_{x}(x_{i})-\mu_{\phi_{x}(x)}\right)\left(\phi_{y}(y_{i})-\mu_{\phi_{y}(y)}\right)^{\top},
\end{align}
where $\mu_{\phi_{x}(x)}=\frac{1}{n}\sum_{i=1}^{n}\phi_{x}(x_{i})$ and $\mu_{\phi_{y}(y)}=\frac{1}{n}\sum_{i=1}^{n}\phi_{y}(y_{i})$ are the means of the two views' kernel mappings, respectively. The form of $\hat{C}_{xx}$ and $\hat{C}_{yy}$ can be obtained analogously.

Let $K_{x}$ denote the kernel matrix such that $K_{x}=H\tilde{K}_{x}H$, where $[\tilde{K}_{x}]_{ij}=k_{x}(x_{i},x_{j})$ and $H=I-\frac{1}{n}{ 1}{ 1}^{\top}$ is a centering matrix, ${1}\in\mathbb{R}^{n}$ being a vector of all ones. And $K_{y}$ is defined similarly. Further, we substitute them into Eq.(\ref{kkcca}) and formulate the objective of KCCA as the following optimization problem:
\begin{align}
&\max_{\alpha,\beta}\frac{\alpha^{\top}K_{x}K_{y}\beta}{\sqrt{\alpha^{T}K_{x}^{2}\alpha\beta^{\top}K_{y}^{2}\beta}}
\end{align}

As discussed in \cite{Hardoon2004}, the above optimization leads to degenerate solutions when either $K_{x}$ or $K_{y}$ is invertible. Thus, we introduce regularization terms and maximize the following regularized expression 
\begin{align}
\max_{\alpha,\beta}\frac{\alpha^{\top}K_{x}K_{y}\beta}{\sqrt{\alpha^{T}\left(K_{x}^{2}+\epsilon_{x}K_{x}\right)\alpha\beta^{\top}\left(K_{y}^{2}+\epsilon_{y}K_{y}\right)\beta}}
\label{kcca1}
\end{align}
Since this new regularized objective function is not affected by re-scaling of $\alpha$ or $\beta$, we assume that the optimization problem is subject to
\begin{align}
\alpha^{\top}K_{x}^{2}\alpha+\epsilon_{x}\alpha^{\top}K_{x}\alpha=1\notag\\
\beta^{\top}K_{y}^{2}\beta+\epsilon_{y}\beta^{\top}K_{y}\beta=1
\label{kcca_constraint}
\end{align}

Similar to the optimized case of CCA, by formulating the Lagrangian dual of Eq.(\ref{kcca1}) with the constraints in Eq.(\ref{kcca_constraint}), it can be shown that the solution to Eq.(\ref{kcca1}) is also equivalent to solving a pair of generalized eigenproblems \cite{Hardoon2004},
\begin{align}
\left(K_{x}+\epsilon_{x}I\right)^{-1}K_{y}\left(K_{y}+\epsilon_{y}I\right)^{-1}K_{x}\alpha=\lambda^{2}\alpha\notag\\
\left(K_{y}+\epsilon_{y}I\right)^{-1}K_{x}\left(K_{x}+\epsilon_{x}I\right)^{-1}K_{y}\beta=\lambda^{2}\beta
\label{kcca_solution}
\end{align} 

Consequently, the statistical properties of KCCA have been investigated from several aspects \cite{BachJ02, Kuss03}. Fukumizu et al. \cite{FukumizuBG07} introduce a mathematical proof of the statistical convergence of KCCA by providing rates for the regularization parameters. Later Hardoon and Shawe-Taylor \cite{HardoonS09} provide a detailed theoretical analysis of KCCA and propose a finite sample statistical analysis of KCCA by using a regression formulation. Cai and Sun \cite{Cai2011} also provide a convergence rate analysis of KCCA under an approximation assumption. However, the problems of choosing appropriate regularization parameters in practice remain largely unsolved.

In addition, KCCA has a closed-form solution via the eigenvalue system in Eq.(\ref{kcca_solution}), but this solution does not scale up to the large size of the training set, due to the problem of time complexity and memory cost. Thus, various approximation methods have been proposed by constructing low-rank approximations of the kernel matrices, including incomplete Cholesky decomposition \cite{BachJ02}, partial Gram-Schmidt orthogonolisation \cite{Hardoon2004}, and block incremental SVD \cite{Brand02,AroraL12}. In addition, the Nystr{\"{o}}m method \cite{WilliamsS00} is widely used to speed up the kernel machines \cite{YangLMJZ12,LeSS13,Lopez-PazSSGS14,WangL15h}. This approach is achieved by carrying out an eigen-decomposition on a lower-dimensional system, and then expanding the results back up to the original dimensions.  

\subsubsection{Deep CCA}
The CCA-like objectives can be naturally applied to neural networks to capture high-level associations between data from multiple views. In the early work, by assuming that different parts of the perceptual input have common causes in the external world, Becker and Hinton \cite{Becker92} present a multilayer nonlinear extension of canonical correlation by maximizing the normalized covariation between the outputs from two neural network modules. Further, Becker \cite{Becker96} explores the idea of maximizing the mutual information between the outputs of different network modules to extract higher order features from coherence inputs.

Later Lai and Fyfe \cite{LaiF98,LaiF99} investigate a neural network implementation of CCA and maximize the correlation (rather than canonical correlation) between the outputs of the networks for different views. Hsieh \cite{Hsieh00} formulates a nonlinear canonical correlation analysis (NLCCA) method using three feedforward neural networks. The first network maximizes the correlation between the canonical variates (the two output neurons), while the remaining two networks map the canonical variates back to the original two sets of variables.

\begin{figure}
\begin{center}
   \includegraphics[scale=0.28]{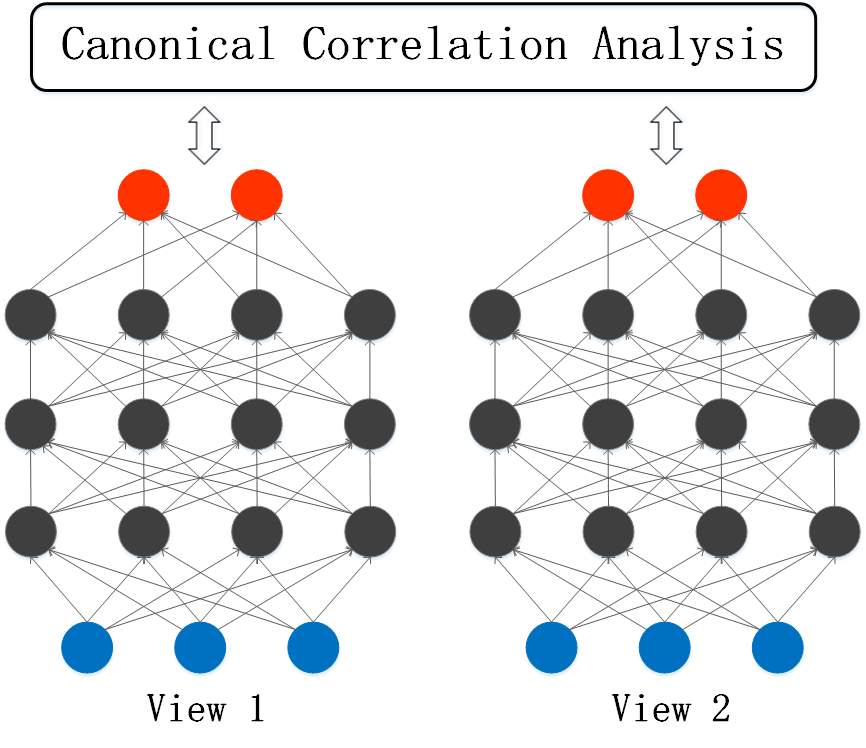}
\end{center}
\vspace*{-0.8em}
\caption{The framework of deep CCA (adapted from \cite{AndrewABL13}), in which the output layers of two deep networks are maximally correlated.}

\label{deepCCA}
\end{figure}

Although multiple CCA-based neural network models have been proposed for decades, the full deep neural network extension of CCA, referred as deep CCA (Figure \ref{deepCCA}), has recently been developed by Andrew et al. \cite{AndrewABL13}. Inspired by the recent success of deep neural networks \cite{salakhutdinov2009deep,krizhevsky2012}, Andrew et al. \cite{AndrewABL13} introduce deep CCA to learn deep nonlinear mappings between two views $\{X,Y\}$ which are maximally correlated. The deep CCA learns representations of the two views by using multiple stacked layers of nonlinear mappings. In particular, assuming for simplicity that a network has $d$ intermediate layers, deep CCA first learns deep representation from $f_{x}(x)=h_{{W}_{x},{b}_{x}}(x)$ with parameters $({{W}_{x}, {b}_{x}})=(W^{1}_{x},b^{1}_{x},W^{2}_{x},b^{2}_{x},\ldots,W^{d}_{x},b^{d}_{x})$, where $W_{x}^{l}(ij)$ denotes the parameters associated with the connection between unit $i$ in layer $l$, and unit $j$ in layer $l+1$. Also, $b_{x}^{l}(j)$ denotes the bias associated with unit $j$ in layer $l+1$. Given a sample of the second view, the representation $f_{y}(y)$ is computed in the same way, with different parameters $\left({W}_{y},{b}_{y}\right)$. The goal of deep CCA is to jointly learn parameters for both views such that $\text{corr}(f_{x}(X),f_{y}(Y))$ is as high as possible. Let $\theta_{x}$ be the vector of all the parameters $({W}^{x},{b}^{x})$ of the first view and similarly for $\theta_{y}$, then 
\begin{align}
(\theta^{*}_{x},\theta^{*}_{y})=\arg\max_{(\theta_{x},\theta_{y})}\text{corr}(f_{x}(X;\theta_{x}),f_{y}(Y;\theta_{y})).
\end{align}
For training deep neural network models, parameters are typically estimated with gradient-based optimization methods. Thus, the parameters $(\theta^{*}_{x},\theta^{*}_{y})$ are also estimated on the training data by following the gradient of the correlation objective, with batch-based algorithms like L-BFGS as in \cite{AndrewABL13} or stochastic optimization with mini-batches \cite{WangALB15,LuWBGL15,WangALS15}.

Deep CCA and its extensions have been widely applied in learning representation tasks in which multiple views of data are provided. Yan and Mikolajczyk \cite{YanM15} learn a joint latent space for matching images and captions with a deep CCA framework, which adopts a GPU implementation and may deal with overfitting. To exploit multilingual context when learning word embeddings, Lu et al. \cite{LuWBGL15} learn deep non-linear embeddings of two languages using the deep CCA. 

Recently, a deep canonically correlated autoencoder (DCCAE) \cite{WangALB15DCCAE} is proposed by combining the advantages of the deep CCA and those of autoencoder-based approaches. In particular, DCCAE consists of two autoencoders and optimizes the combination of canonical correlation between the learned bottleneck representations and the reconstruction errors of the autoencoders. This optimization offers a trade-off between information captured in the embedding within each view on one aspect, and the information in the relationship across views on the other.

\subsection{Distance and Similarity-based Alignment}
In this section we will review the multi-view representation learning techniques from the perspective of distance and similarity-based alignment: partial least squares, cross-modal ranking, cross-modal hashing, and deep cross-view embedding models.

\subsubsection{Partial Least Squares}
Partial Least Squares (PLS) \cite{wold1982soft,Rosipal2006,SchwartzKHD09} is a wide class of methods for modeling relations between sets of observed variables. It has been a popular tool for regression and classification as well  as dimensionality reduction, especially in the field of chemometrics \cite{Wold2001,barkerpls}. The underlying assumption of all PLS methods is that the observed data are generated by a process which is driven by a small number of latent variables. In particular, PLS creates orthogonal latent vectors by maximizing the covariance between different sets of variables.
  
Given a pair of datasets $X=[x_{1},\ldots,x_{n}]\in\mathbb{R}^{d_{x}\times n}$ and $Y=[y_{1},\ldots,y_{n}]\in\mathbb{R}^{d_{y}\times n}$, a $k$-dimensional PLS solution can be parameterized by a pair of matrices $W_{x}\in\mathbb{R}^{d_{x}\times k}$ and $W_{y}\in\mathbb{R}^{d_{y}\times k}$ \cite{AroraCLS12}. The PLS problem can now be expressed as: 
\begin{align}
\max_{W_{x},W_{y}} &\text{tr}\left(W_{x}^{\top}C_{xy}W_{y}\right)\notag\\
\text{s.t.} ~~~~~&W_{x}^{\top}W_{x}=I, W_{y}^{\top}W_{y}=I.
\end{align} 
It can be shown that the columns of the optimal $W_{x}$ and $W_{y}$ correspond to the singular vectors of covariance matrix $C_{xy}=\mathbb{E}[xy^{\top}]$. Like the CCA objective, PLS is also an optimization of an expectation subject to fixed constraints. 

In essence, CCA finds the directions of maximum correlation while PLS finds the directions of maximum covariance. Covariance and correlation are two different statistical measures for describing how variables covary. It has been shown that there are close connections between PLS and CCA in several aspects \cite{barkerpls,Rosipal2006}. Guo and Mu \cite{GuoM13} investigate the CCA based methods, including linear CCA, regularized CCA, and kernel CCA, and compare them with the PLS models in solving the joint estimation problem. In particular, they provide a consistent ranking of the above methods in estimating age, gender, and ethnicity.  

Further, Li et al. \cite{Li2003MCP} introduce a least square form of PLS, called cross-modal factor analysis (CFA). CFA aims to find orthogonal transformation matrices $W_{x}$ and $W_{y}$ by minimizing the following expression:
\begin{align}
&||X^{\top}W_{x}-Y^{\top}W_{y}||^{2}_{F} \notag\\
\text{subject to:} &~~~W_{x}^{\top}W_{x}=I, ~~~W_{y}^{\top}W_{y}=I 
\label{CFA}
\end{align}
where $||\cdot||_{F}$ denotes the Frobenius norm. It can be easily verified that the above optimization problem in Eq.(\ref{CFA}) has the same solution as that to PLS. Several extensions of CFA are presented by incorporating non-linearity and supervised information \cite{WangGV12,PereiraCDRLLV14,WangWTDZC15}.

\subsubsection{Cross-Modal Ranking}
Motivated by incorporating ranking information into multi-modal embedding learning, cross-modal ranking has attracted much attention in the literature \cite{BaiWGCSQCW10,WestonBU10,GrangierB08}. Bai et al. \cite{BaiWGCSQCW10} present a supervised semantic indexing (SSI) model which defines a class of non-linear models that are discriminatively trained to map multi-modal input pairs into ranking scores.

In particular, SSI attempts to learn a similarity function $f(q,d)$ between a text query $q$ and an image $d$ according to a pre-defined ranking loss. The learned function $f$ directly maps each text-image pair to a ranking score based on their semantic relevance. Given a text query $q\in\mathbb{R}^{m}$ and an image $d\in\mathbb{R}^{n}$, SSI intends to find a linear scoring function to measure the relevance of $d$ given $q$:
\begin{align}
f(q,d)=q^{\top}Wd=\sum_{i=1}^{m}\sum_{j=1}^{n}q_{i}W_{ij}d_{j}
\label{SSI}
\end{align}
where $f(q,d)$ is defined as the score between the query $q$ and the image $d$, and the parameter $W\in\mathbb{R}^{m\times n}$ captures the correspondence between the two different modalities of the data: $W_{ij}$ represents the correlation between the $i$-th dimension of the text space and the $j$-th dimension of the image space. Note that this way of embedding allows both positive and negative correlations between different modalities since both positive and negative values are allowed in $W_{ij}$.

Given the similarity function in Eq.(\ref{SSI}) and a set of tuples, where each tuple contains a query $q$, a relevant image $d^{+}$ and an irrelevant image $d^{-}$, SSI attempts to choose the scoring function $f(q,d)$ such that $f(q,d^{+})>f(q,d^{-})$, expressing that $d^{+}$ should be ranked higher than $d^{-}$. For this purpose, SSI exploits the margin ranking loss \cite{herbrich00ordinal} which has already been widely used in information retrieval, and minimizes:
\begin{align}
\sum_{(q,d^{+},d^{-})} \max(0,1-q^{T}Wd^{+}+q^{T}Wd^{-})
\end{align}
This optimization problem can be solved through stochastic gradient descent \cite{BurgesSRLDHH05},
\begin{align}
W\leftarrow &W+\lambda\left(q(d^{+})^{\top}-q(d^{-})^{\top}\right), \notag\\
&~~~~~~~~~~~~~~~~~~~~~\text{if} ~~~1-q^{T}Wd^{+}+q^{T}Wd^{-}>0 
\end{align}

In fact, this method is a special margin ranking perceptron \cite{Collins2002}, which has been shown to be equivalent to SVM \cite{CollobertB04}. In contrast to classical SVM, stochastic training is highly scalable and is easy to implement for millions of training examples. However, dealing with the models on all the pairs of multi-modalities input features are still computationally challenging. Thus, SSI also proposes several improvements to the above basic model for addressing this issue, including low-rank representation, sparsification, and correlated feature hashing. For more detailed information, please refer to \cite{BaiWGCSQCW10}.

Further, to exploit the advantage of online learning of kernel-based classifiers, Grangier and Bengio \cite{GrangierB08} propose a discriminative cross-modal ranking model called Passive-Aggressive Model for Image Retrieval (PAMIR), which not only adopts a learning criterion related to the final retrieval performance, but also considers different image kernels.

\subsubsection{Cross-Modal Hashing}
To speed up the cross-modal similarity search, a variety of multi-modal hashing methods have been gradually proposed \cite{BronsteinBMP10,KumarU11,ZhenY12NIPS,ZhuHSZ13,ZhenGYZL16}. The principle of the multi-modal hashing methods is to map the high dimensional multi-modal data into a common hash code so that similar cross-modal data objects have the same or similar hash codes.

Bronstein et al. \cite{BronsteinBMP10} propose a hashing-based model, called cross-modal similarity sensitive hashing (CMSSH), which approaches the cross-modality similarity learning problem by embedding the multi-modal data into a common metric space. The similarity is parameterized by the embedding itself. The goal of cross-modality similarity learning is to construct the similarity function between points from different spaces, $X\in\mathbb{R}^{d_{1}}$ and $Y\in\mathbb{R}^{d_{2}}$. Assume that the unknown binary similarity function is $s: X\times Y\rightarrow \{\pm1\}$; the classical cross-modality similarity learning aims at finding a binary similarity function $\hat{s}$ on $X\times Y$ approximating $s$. Recent work attempts to solve the problem of cross-modality similarity leaning as an multi-view representation learning problem.

In particular, CMSSH proposes to construct two maps: $\xi:X\rightarrow \mathbb{H}^{n}$ and $\eta:Y\rightarrow \mathbb{H}^{n}$, where $\mathbb{H}^{n}$ denotes the $n$-dimensional Hamming space. Such mappings encode the multi-modal data into two $n$-bit binary strings so that $d_{\mathbb{H}^{n}}(\xi(x),\eta(y))$ is small for $s(x,y)=+1$ and large for $s(x,y)=-1$ with a high probability. Consequently, this hamming embedding can be interpreted as cross-modal similarity-sensitive hashing, under which positive pairs have a high collision probability, while negative pairs are unlikely to collide. Such hashing also acts as a way of multi-modal dimensionality reduction when $d_{1},d_{2}\gg n$.

The $n$-dimensional Hamming embedding for $X$ can be considered as a vector $\xi(x)=(\xi_{1}(x),\ldots,\xi_{n}(x))$ of binary embeddings of the form
\begin{align}
\xi_{i}(x) = \left\{ \begin{array}{ll}0 & \mbox{if  } f_{i}(x)\leq 0,\\
1 & \mbox{if  } f_{i}(x)>0,\end{array} \right.
\end{align}
parameterized by a projection $f_{i}:X\rightarrow \mathbb{R}$. Similarly, $\eta_{i}$ is a binary mapping parameterized by projection $g_{i}:Y\rightarrow \mathbb{R}$. 

Following the greedy approach \cite{Shakhnarovich2005}, the Hamming metric can be constructed sequentially as a superposition of weak binary classifiers on pairs of data points, 
\begin{align}
h_{i}(x,y) &= \left\{ \begin{array}{ll}+1 & \mbox{if  } \xi_{i}(x)=\eta_{i}(y),\\
-1 & \mbox{otherwise },\end{array} \right.\notag\\
&=(2 \xi_{i}(x)-1)(2\eta_{i}(y)-1),
\end{align}
Here, a simple strategy for the mappings is affine projection, such as $f_{i}(x)=p_{i}^{\top}+a_{i}$ and $g_{i}(y)=q_{i}^{\top}y+b_{i}$. It can be extended to complex projections easily. 

Observing the resemblance to sequentially binary classifiers, boosted cross-modality similarity learning algorithms are introduced based on the standard AdaBoost procedure \cite{FreundS97}. CMSSH has shown its utility and efficiency in several multi-view learning applications including cross-representation shape retrieval and alignment of multi-modal medical images.

However, CMSSH only considers the inter-view correlation but ignores the intra-view similarity \cite{YuWYTLZ14}. Kumar and Udupa \cite{KumarU11} extend Spectral Hashing \cite{WeissTF08} from the single view setting to the multi-view scenario and present cross view hashing (CVH), which attempts to find hash functions that map similar objects to similar codewords over all the views so that inter-view and intra-view similarities are both preserved. Gong and Lazebink \cite{GongL11} combine iterative quantization with CCA to exploit cross-modal embeddings for learning similarity preserving binary codes. Consequently, Zhen and Yang \cite{ZhenY12NIPS} present co-regularized hashing (CRH) for multi-modal data based on a boosted co-regularization framework. The hash functions for each bit of the hash codes are learned by solving DC (difference of convex functions) programs, while the learning for multiple bits is performed via a boosting procedure. Later Song et al. \cite{SongYYHS13} introduce an inter-media hashing (IMH) model by jointly capturing inter-media and intra-media consistency.
 
\subsubsection{Deep Cross-View Embedding Models}
\begin{figure}
\begin{center}
   \includegraphics[scale=0.35]{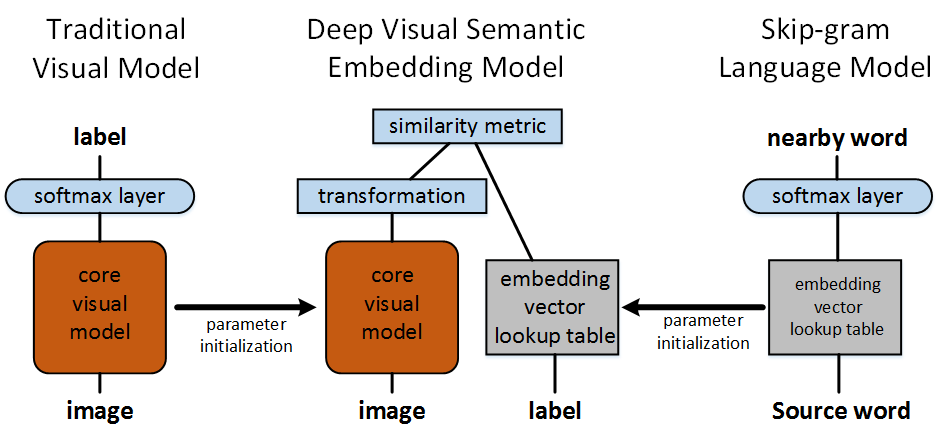}
\end{center}
\caption{The DeViSE model (adapted from \cite{frome2013devise}), which is initialized with parameters pre-trained at the lower layers of the visual object categorization network and the skip-gram language model.}
\label{DeViSE}
\end{figure}

Deep cross-view embedding models have become increasingly popular in the applications including cross-media retrieval \cite{frome2013devise,HuangHGDAH13,ElkahkySH15,dong2016word2visualvec} and multi-modal distributional semantic learning \cite{kiros2014multimodal,lazaridou2015combining}. Frome et al. \cite{frome2013devise} propose a deep visual-semantic embedding model (DeViSE), which connects two deep neural networks by a cross-modal mapping. As shown in Figure {\ref{DeViSE}}, DeViSE is first initialized with a pre-trained neural network language model \cite{mikolov2013efficient} and a pre-trained deep visual-semantic model \cite{krizhevsky2012}. Consequently, a linear transformation is exploited to map the representation at the top of the core visual model into the learned dense vector representations by the neural language model. 

Following the setup of loss function in \cite{BaiWGCSQCW10}, DeViSE employs a combination of dot-product similarity and hinge rank loss so that the model has the ability of producing a higher dot-product similarity between the visual model output and the vector representation of the correct label than between the visual output and the other randomly chosen text terms. The per-training example hinge rank loss is defined as follows:
\begin{align}
\sum_{j\neq label}\max\left[0,\text{margin}-\vec{t}_{\text{label}}M\vec{v}{(\text{image})}+\vec{t}_{j}M\vec{v}{(\text{image})}\right]
\end{align}
where $\vec{v}(\text{image})$ is a column vector denoting the output of the top layer of the core visual network for the given image, $M$ is the mapping matrix of the linear transformation layer, $\vec{t}_{\text{label}}$ is a row vector denoting the learned embedding vector for the provided text label, and $\vec{t}_{j}$ are the embeddings of the other text terms.
This DeViSE model is trained by asynchronous stochastic gradient descent on a distributed computing platform \cite{dean2012large}.

Inspired by the success of DeViSE, Norouzi et al. \cite{norouzi2013zero} propose a convex combination of semantic embedding model (ConSE) for mapping images into continuous semantic embedding spaces. Unlike DeViSE, this ConSE model keeps the softmax layer of the convolutional net intact. Given a test image, ConSE simply runs the convolutional classifier and considers the convex combination of the semantic embedding vectors from the top $T$ predictions as its corresponding semantic embedding vector. Further, Fang et al. \cite{fang2015captions} develop a deep multi-modal similarity model that learns two neural networks to map image and text fragments to a common vector representation.

In addition, Xu et al. \cite{xu2015jointly} propose a unified framework that jointly models video and the corresponding text sentence. In this joint architecture, the goal is to learn a function $f(\mathcal{V}):\mathcal{V}\rightarrow \mathcal{T}$, where $\mathcal{V}$ represents the low-level features extracted from video, and $\mathcal{T}$ is the high-level text description of the video. A joint model $\mathcal{P}$ is designed to connect these two levels of information. It consists of three parts: compositional language model $M_{L}:T\rightarrow T_{f}$, deep video model $M_{V}:V\rightarrow V_{f}$, and a joint embedding model $E(V_{f},T_{f})$, such that
\begin{align}
\mathcal{P}: M_{V}(V)\longrightarrow V_{f}\leftrightarrow E(V_{f},T_{f})\leftrightarrow T_{f}\longleftarrow M_{L}(T)
\end{align}
where $V_{f}$ and $T_{f}$ are the outputs of the deep video model and the compositional language model, respectively. In this joint embedding model, the distance between the outputs of the deep video model and those of the compositional language model in the joint space is minimized to make them alignment.

\section{Multi-View Representation Fusion}
Multi-view representation fusion methods aims to integrate multi-view inputs into a single compact representation. Representative examples can be reviewed from two perspectives: 1) graphical models; 2) neural network-based models. In this section we first review the generative multi-view representation learning techniques. Then typical examples of neural network-based fusion methods are surveyed to demonstrate the expressive power of the deep multi-view joint representation.
 
\subsection{Graphical Model-based Representation Fusion}
In this section we will review the multi-view representation learning techniques from the generative perspective: multi-modal latent Dirichlet allocation, multi-view sparse coding, multi-view latent space Markov networks, and multi-modal deep Boltzmann machine.
   
\subsubsection{Multi-Modal Latent Dirichlet Allocation}
Latent Dirichlet allocation (LDA) \cite{BleiNJ03} is a generative probabilistic model for collections of a corpus. It proceeds beyond PLSA through providing a generative model at word and document levels simultaneously. In particular, LDA is a three-level hierarchical Bayesian network that models each document as a finite mixture over an underlying set of topics.


\begin{figure}
\begin{center}
   \includegraphics[scale=0.35]{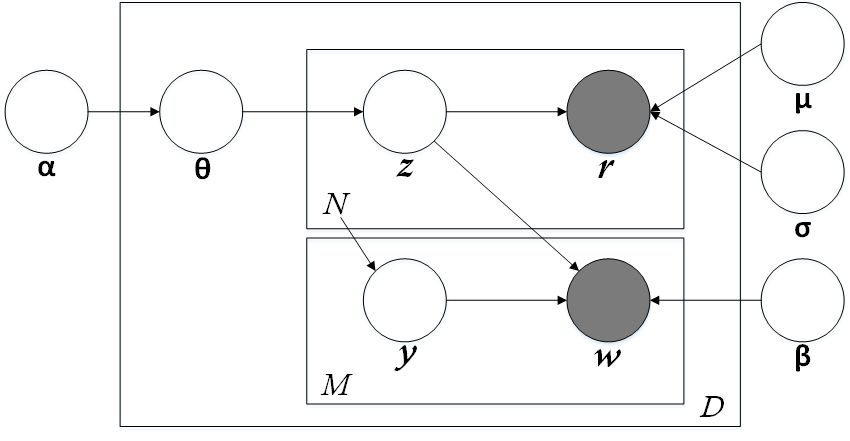}
\end{center}
\caption{The graphical model representation of Corr-LDA model (adapted from \cite{BleiJ03}).}
\label{Corr-LDA}
\end{figure}

As a generative model, LDA is extendable to multi-view data. Blei and Jordan \cite{BleiJ03} propose a correspondence LDA (Corr-LDA) model, which not only allows simultaneous dimensionality reduction in the representation of region descriptions and words, but also models the conditional correspondence between their respectively reduced representations. The graphical model of Corr-LDA is depicted in Figure \ref{Corr-LDA}. This model can be viewed in terms of a generative process that first generates the region descriptions and subsequently generates the caption words.

In particular, let $\mathbf{z}=\{z_{1},z_{2},\ldots,z_{N}\}$ be the latent variables that generate the image, and let $\mathbf{y}=\{y_{1},y_{2},\ldots,y_{M}\}$ be discrete indexing variables that take values from $1$ to $N$ with an equal probability. Each image and its corresponding caption are represented as a pair $(\mathbf{r,w})$. The first element $\mathbf{r}=\{r_{1},\ldots,r_{N}\}$ is a collection of $N$ feature vectors associated with the regions of an image. The second element $\mathbf{w}=\{w_{1},\ldots,w_{M}\}$ is the collection of $M$ words of the caption. Given $N$ and $M$, a $K$-factor Corr-LDA model assumes the following generative process for an image-caption pair $(\mathbf{r,w})$:
\begin{enumerate}
\item Sample $\theta\sim\text{Dir}(\theta|\alpha)$.
\item For each image region $r_{n}$, $n\in\{1,\ldots,N\}$:
\begin{enumerate}
\item Sample $z_{n}\sim\text{Mult}(\theta)$
\item Sample $r_{n}\sim p(r|z_{n},\mu,\sigma)$ from a multivariate Gaussian distribution conditioned on $z_{n}$.
\end{enumerate}
\item For each caption word $w_{m}$, $m\in\{1,\ldots,M\}$:
\begin{enumerate}
\item Sample $y_{m}\sim \text{Unif}(1,\ldots,N)$
\item Sample $w_{m}\sim p(w|y_{m},{\mathbf z},\beta)$ from a multinomial distribution conditioned on the $z_{y_{m}}$ factor.
\end{enumerate}
\end{enumerate}

Consequently, Corr-LDA specifies the following joint distribution on image regions, caption words, and latent variables:
\begin{align}
p(\mathbf{r,w,}\theta,\mathbf{z,y})=p(\theta|\alpha)&\left(\prod_{n=1}^{N}p(z_{n}|\theta)p(r_{n}|z_{n},\mu,\sigma)\right)\notag\\
&\cdot\left(\prod_{m=1}^{M}p(y_{m}|N)p(w_{m}|y_{m},\mathbf{z},\beta)\right)
\end{align} 
Note that exact probabilistic inference for Corr-LDA is intractable and we employ variational inference methods to approximate the posterior distribution over the latent variables given a particular pair of image-caption.

Further, supervised multi-modal LDA models are subsequently proposed to make effective use of the discriminative information. For instance, Wang et al. \cite{WangBL09} develop a multi-modal probabilistic model for jointly modeling an image, its class label, and its annotations, called multi-class supervised LDA with annotations, which treats the class label as a global description of the image, and treats the annotation terms as local descriptions of parts of the image. Cao et al. \cite{CaoF07} propose a spatially coherent latent topic model (Spatial-LTM), which represents an image containing objects in two different modalities: appearance features and salient image patches.

\subsubsection{Multi-View Sparse Coding}
Multi-view sparse coding \cite{JiaSD10,CaoJMPCN13,liu2014multiview} relates a shared latent representation to the multi-view data through a set of linear mappings, which we define as the {\em dictionaries}. It has the property of finding shared representation $h^{*}$ which selects the most appropriate bases and zeros the others, resulting in a high degree of correlation with the multi-view input. This property is owing to the explaining away effect which aries naturally in directed graphical models \cite{PRMC}.

Given a pair of datasets $\{X,Y\}$, a non-probabilistic multi-view sparse coding scheme can be formulated as learning the representation or code vector with respect to a multi-view sample:
\begin{align}
h^{*}=\arg \min_{h} \|x-W_{x}h\|_{2}^{2}+\|y-W_{y}h\|_{2}^{2}+\lambda \|h\|_{1}
\label{msc}
\end{align}
Learning the pair of dictionaries $\{W_{x},W_{y}\}$ can be implemented by optimizing the following objective with respect to $W_{x}$ and $W_{y}$:
\begin{align}
\mathcal{J}_{W_{x},W_{y}}=\sum_{i}\left(\|x_{i}-W_{x}h^{*}_{i}\|_{2}^{2}+\|y_{i}-W_{y}h^{*}_{i}\|_{2}^{2}\right)
\label{optimized_W}
\end{align}
where $x_{i}$ and $y_{i}$ are the two modal inputs and $h^{*}$ is the corresponding shared sparse representation computed with Eq.(\ref{msc}). In particular, $W_{x}$ and $W_{y}$ are usually regularized by the constraint of having unit-norm columns.

The above regularized form of multi-view sparse coding can be generalized as a probabilistic model. In probabilistic multi-view sparse coding, we assume the following generative distributions,
\begin{align}
p(h)&=\prod_{j}^{d_h} \frac{\lambda}{2}\text{exp}(-\lambda |h_{j}|)\notag \\
\forall_{i=1}^{n}: ~~~~~~&p(x_{i}|h)=\mathcal{N}(x_{i};W_{x}h+\mu_{x_{i}},\sigma^{2}_{x_{i}}{\bold{I}})\notag\\
&p(y_{i}|h)=\mathcal{N}(y_{i};W_{y}h+\mu_{y_{i}},\sigma^{2}_{y_{i}}{\bold{I}})    
\end{align}
In this case of multi-view probabilistic sparse coding, we aims to obtain a sparse multi-view representation by computing the MAP (maximum {\em a posteriori}) value of $h$: i.e., $h^{*}=\arg\max_{h}p(h|x,y)$ instead of its expected value $E[h|x,y]$. From this viewpoint, learning parameters $W_{x}$ and $W_{y}$ can be accomplished by maximizing the likelihood of the data given the joint MAP values of $h^{*}$: $\arg\max_{W_{x},W_{y}}\prod_{i}p(x_{i}|h^{*})p(y_{i}|h^{*})$. Generally, expectation-maximization can be exploited to learn dictionaries $\{W_x,W_{y}\}$ and shared representation $h^{*}$ alternately.

One might expect that multi-view sparse representation would significantly leverage the performance especially when features for different views are complementary to one another and indeed it seems to be the case. There are numerous examples of its successful applications as a multi-view feature learning scheme, including human pose estimation \cite{JiaSD10}, image classification \cite{HanWTSZJ12}, web data mining \cite{YuRT14}, as well as cross-media retrieval \cite{ZhuangWWZL13, WuYYTZZ14}. For example, Liu et al. \cite{liu2014multiview} introduce multi-view Hessian discriminative sparse coding (mHDSC) which combines multi-view discriminative sparse coding with Hessian regularization. mHDSC can exploit the local geometry of the data distribution based on the Hessian regularization and fully takes advantage of the complementary information of multiview data to improve the learning performance. 

\subsubsection{Multi-View Latent Space Markov Networks}
Undirected graphical models, also called Markov random fields, have many special cases, including the exponential family Harmonium \cite{WellingRH04} and restricted Boltzmann machine \cite{Hinton02}. Within the context of unsupervised multi-view feature learning, Xing et al. \cite{XingYH05} first introduce a particular form of multi-view latent space Markov network model called multi-wing harmonium model. This model can be viewed as an undirected counterpart of the aforementioned directed aspect models such as multi-modal LDA \cite{BleiJ03}, with the advantages that inference is fast due to the conditional independence of the hidden units and that topic mixing can be achieved by document- and feature-specific combination of aspects.

For simplicity, we begin with dual-wing harmonium model, which consists of two modalities of input units ${X}=\{x_{i}\}_{i=1}^{n}$, ${Y}=\{y_{j}\}_{j=1}^{n}$, and a set of hidden units ${H}=\{h_{k}\}_{k=1}^{n}$. In this dual-wing harmonium, each modality of input units and the hidden units constructs a complete bipartite graph where units in the same set have no connections but are fully connected to units in the other set. In addition, there are no connections between two input modalities. In particular, consider all the cases where all the observed and hidden variables are from exponential family; we have
\begin{align}
p(x_{i})=&\text{exp}\{\theta_{i}^{T}\phi(x_{i})-A(\theta_{i})\}\notag\\
p(y_{j})=&\text{exp}\{\eta_{j}^{T}\psi(y_{j})-B(\eta_{j})\}\notag\\
p(h_{k})=&\text{exp}\{\lambda_{k}^{T}\varphi(h_{k})-C(\lambda_{k})\}
\end{align}
where $\phi(\cdot)$, $\psi(\cdot)$, and $\varphi(\cdot)$ are potentials over cliques formed by individual nodes, $\theta_{i}$, $\eta_{j}$, and $\lambda_{k}$ are the associated weights of potential functions, and $A(\cdot)$, $B(\cdot)$, and $C(\cdot)$ are log partition functions.

Through coupling the random variables in the log-domain and introducing other additional terms, we obtain the joint distribution $p(X,Y,H)$ as follows:
{\small \begin{align}
p({X,Y,H})\propto ~~&\text{exp}\big{\{} \sum_{i}\theta_{i}^{T}\phi(x_{i})+\sum_{j}\eta_{j}^{T} \psi(y_{j})+\sum_{k}\lambda_{k}^{T}\varphi(h_{k})\notag
\\
+&\sum_{ik}\phi(x_{i})^{T}W_{ik}\varphi(h_{k})+\sum_{jk} \psi (y_{j})^{T}U_{jk}\varphi(h_{k})\big{\}}
\label{MMN}
\end{align}}
where $\phi(x_{i})\varphi(h_{k})$, $\psi(y_{j})\varphi(h_{k})$ are potentials over cliques consisting of pairwise linked nodes, and $W_{ik}$, $U_{jk}$ are the associated weights of potential functions. From the joint distribution, we can derive the conditional distributions
\begin{align}
p(x_{i}|H)&\propto \text{exp}\{\tilde{\theta}^{T}_{i}\phi(x_{i})-A(\tilde{\theta}_{i})\}\notag\\
p(y_{j}|H)&\propto \text{exp}\{\tilde{\eta}^{T}_{j}\psi(y_{j})-B(\tilde{\eta}_{j})\}\notag \\
p(h_{k}|X,Y)&\propto \text{exp}\{\tilde{\lambda}_{k}^{T}\varphi(h_{k})-C(\tilde{\lambda}_{k})\}
\end{align}
where the shifted parameters $\tilde{\theta}_{i}=\theta_{i}+\sum_{k}W_{ik}\varphi(h_{k})$, $\tilde{\eta}_{j}=\eta_{j}+\sum_{k}U_{jk}\varphi(h_{k})$, and $\tilde{\lambda}_{k}=\lambda_{k}+\sum_{i}W_{ik}\phi(x_{i})+\sum_{j}U_{jk}\psi(y_{j})$.

In training probabilistic models parameters are typically updated in order to maximize the {\em likelihood of the training data}. The updating rules can be obtained by taking derivative of the log-likelihood of the sample defined in Eq.(\ref{MMN}) with respect to the model parameters. The multi-wing model can be directly obtained by extending the dual-wing model when the multi-modal input data are observed.

Further, Chen et al. \cite{ChenZX10} present a multi-view latent space Markov network and its large-margin extension that satisfies a weak conditional independence assumption that data from different views and the response variables are conditionally independent given a set of latent variables. In addition, Xie and Xing \cite{XieX13} propose a multi-modal distance metric learning (MMDML) framework based on the multi-wing harmonium model and metric learning method by \cite{XingNJR02}. This MMDML provides a principled way to embed data of arbitrary modalities into a single latent space where distance supervision is leveraged.

\subsubsection{Multi-Modal Deep Boltzmann Machine}
Restricted Boltzmann Machines (RBM) \cite{Rumelhart1986} is an undirected graphical model that can learn the distribution of training data. The model consists of stochastically visible units ${\bf v}\in\{0,1\}^{d_{v}}$ and stochastically hidden units ${\bf h}\in\{0,1\}^{d_{h}}$, which seeks to minimize the following energy function $E:\{0,1\}^{d_{v}+d_{h}}\rightarrow \mathbb{R}:$
\begin{align}
E({\bf v,h;} ~\theta)=-\sum_{i=1}^{d_{v}}\sum_{j=1}^{d_{h}}v_{i}W_{ij}h_{j}-\sum_{i=1}^{d_{v}}b_{i}v_{i}-\sum_{j=1}^{d_{h}}a_{j}h_{j}
\end{align}
where $\theta=\{{\bf a,b,}W\}$ are the model parameters. Consequently, the joint distribution over the visible and hidden units is defined by:
\begin{align}
P({\bf v,h;} ~\theta)=\frac{1}{\mathcal{Z}(\theta)}\text{exp}\left(-E({\bf v,h;}~\theta)\right).
\end{align}
When considering modeling visible real-valued or sparse count data, this RBM can be easily extended to corresponding variants, e.g., Gaussian RBM \cite{WellingRH04} and replicated softmax RBM \cite{hinton2009replicated}.

\begin{figure}
\begin{center}
   \includegraphics[height=4.3cm,width=7.5cm]{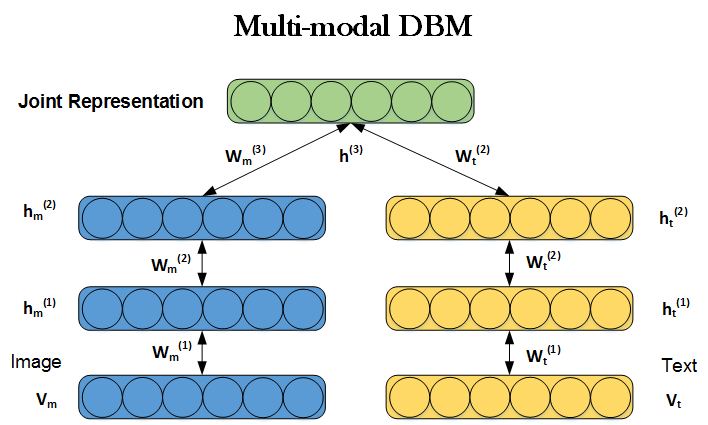}
\end{center}
\caption{The graphical model of deep multi-modal RBM (adapted from \cite{SrivastavaS12}), which captures the joint distribution over image and text inputs.}
\label{MRBM}
\end{figure}

A deep Boltzmann machine (DBM) is a generative network of stochastic binary units. It consists of a set of visible units ${\bf v}\in\{0,1\}^{d_{v}}$, and a sequence of layers of hidden units ${\bf h}^{(1)}\in\{0,1\}^{d_{h_1}},{\bf h}^{(2)}\in\{0,1\}^{d_{h_2}},\ldots,{\bf h}^{(L)}\in\{0,1\}^{d_{h_L}}$. Here connections between hidden units are only allowed in adjacent layers.
Let us take a DBM with two hidden layers for example. By ignoring bias terms, the energy of the joint configuration $\{\bf{v,h}\}$ is defined as 
\begin{align}
E({\bf v,h;}~\theta)=-{\bf v}^{\top}{ W}^{(1)}{\bf h}^{(1)}-{\bf h}^{(1)}{W}^{(2)}{\bf h}^{(2)}
\end{align}
where ${\bf h}=\{\bf h^{(1)},h^{(2)}\}$ represents the set of hidden units, and $\theta=\{ W^{(1)}, W^{(2)}\}$ are the model parameters that denote visible-to-hidden and hidden-to-hidden symmetric interaction terms. Further, this binary-to-binary DBM can also be easily extended to modeling dense real-valued or sparse count data.

By extending the setup of the DBM, Srivastava and Salakhutdinov \cite{SrivastavaS12} propose a deep multi-modal RBM to model the relationship between imagery and text. In particular, each data modality is modeled using a separate two-layer DBM and then an additional layer of binary hidden units on top of them is added to learn the shared representation. 

Let ${\bf v_{m}\in\mathbb{R}^{d_{v_m}}}$ denote an image input and ${\bf v_{t}\in\mathbb{R}^{d_{v_t}}}$ denote a text input. By ignoring bias terms on the hidden units for clarity, the distribution of $\bf v_{m}$ in the image-specific two-layer DBM is given as follows:
\begin{align}
&P({\bf v_{m}};~\theta)=\sum_{\bf h^{(1)},h^{(2)}}P({\bf v}_{m},{\bf h^{(1)}},{\bf h^{(2)}};~\theta)\notag\\
&=\frac{1}{\mathcal{Z}(\theta)}\sum_{\bf h^{(1)}, h^{(2)}}\text{exp}\Big(-\sum_{i=1}^{d_{v_m}}\frac{(v_{mi}-b_{i})^{2}}{2\sigma^{2}_{i}}+\sum_{i=1}^{d_{v_m}}\sum_{j=1}^{d_{h_1}}\frac{v_{mi}}{\sigma_{i}}W_{ij}^{(1)}\notag\\
&~~~~~~~~~~~~~~~~~~~~~~~~~~~~~~~~~~~~~+\sum_{j=1}^{d_{h_1}}\sum_{l=1}^{d_{h_2}}h_{j}^{(1)}W_{jl}^{(2)}h_{l}^{(2)}\Big).
\end{align}
Similarly, the text-specific two-layer DBM can also be defined by combining a replicated softmax model with a binary RBM.

Consequently, the deep multi-modal DBM has been presented by combining the image-specific and text-specific two-layer DBM with an additional layer of binary hidden units on top of them. The particular graphical model is shown in Figure \ref{MRBM}. The joint distribution over the multi-modal input can be written as:
\begin{align}
P({\bf v}^{m},{\bf v}^{t};\theta)=&\sum_{{\bf h}_{m}^{(2)},{\bf h}_{t}^{(2)},{\bf h}^{(3)}}P\big({\bf h}_{m}^{(2)},{\bf h}_{t}^{(2)},{\bf h}^{(3)}\big)\Big(\sum_{{\bf h}_{m}^{(1)}}P\big({\bf v}_{m},\notag\\
&{\bf h}_{m}^{(1)},{\bf h}_{m}^{(2)}\big)\Big)\Big(\sum_{{\bf h}_{t}^{(1)}}P\big({\bf v}_{t},{\bf h}_{t}^{(1)},{\bf h}_{t}^{(2)}\big)\Big)
\end{align}
Like RBM, exact maximum likelihood learning in this model is also intractable, while efficient approximate learning can be implemented by using mean-field inference to estimate data-dependent expectations, and an MCMC based stochastic approximation procedure to approximate the model's expected sufficient statistics \cite{salakhutdinov2009deep}.

Multi-modal DBM has been widely used for multi-view representation learning \cite{huang2013audio,hu2013multimodal,ge2013multi}. Hu et al. \cite{hu2013multimodal} employ the multi-modal DBM to learn joint representation for predicting answers in cQA portal. Ge et al. \cite{ge2013multi} apply the multi-modal RBM to determining information trustworthiness, in which the learned joint representation denotes the consistent latent reasons that underline users' ratings from multiple sources. Pang and Ngo \cite{pang2015mutlimodal} propose to learn a joint density model for emotion prediction in user-generated videos with a deep multi-modal Boltzmann machine. This multi-modal DBM is exploited to model the joint distribution over visual, auditory, and textual features. Here Gaussian RBM is used to model the distributions over the visual and auditory features, and replicated softmax topic model is applied for mining the textual features.

\subsection{Neural Network-based Representation Fusion}
In this section we will review the multi-view representation learning techniques from the neural network perspective: multi-modal deep autoencoders, multi-view convolutional neural network, and multi-modal recurrent neural network.
\subsubsection{Multi-Modal Deep Autoencoder}
An autoencoder is an unsupervised neural network which learns latent representation through input reconstruction \cite{bengio2013representation}. It consists of an encoder $f_{\theta}(\cdot)$ which allows an efficient computation of a latent feature $h=f_{\theta}(x)$ from an input $x$. A decoder $g_{\theta'}(\cdot)$ then aims to map from feature back into the reconstructed input, $\hat{x}=g_{\theta'}(h)$. Consequently, basic auto-encoder training seeks to minimize the following reconstruction error,      
\begin{align}
\mathcal{J}_{AE}\left(\theta, \acute{\theta}\right)=\sum_{i}L\left(x^{i},g_{\theta'}(f_{\theta}(x^{i}))\right)
\end{align}
where $\theta$ and $\theta'$ are parameters of the encoder and decoder and are usually optimized by stochastic gradient descent. Similar to the setup of multi-layer perceptron, the encoder and decoder usually adopt affine mappings, optionally followed by a non-linearity activation:
\begin{align}
&f_{\theta}(x)=s_{f}\left(Wx+b\right)\notag\\
&g_{\theta'}(h)=s_{g}\left(W'h+b'\right)
\end{align}
where $\{W,b,W',b'\}$ are the set of parameters of the network and $s_f$ and $s_g$ are the activation functions of the encoder and decoder, such as element-wise sigmoid and hyperbolic tangent functions. Further, denoising autoencoder is introduced to learn more stable and robust representation by reconstructing a clean input $x^{i}$ from a corrupted version $\tilde{x}^{i}$ and its reconstruction objective function is as follows:
\begin{align}
\mathcal{J}_{DAE}\left(\theta, \acute{\theta}\right)=\sum_{i}L\left(x^{i},g_{\theta'}\left(f_{\theta}\left(\tilde{x}^{i}\right)\right)\right)
\end{align}
where Gaussian noise and masking noise are very common options for corruption process.

\begin{figure}
\begin{center}
   \includegraphics[height=4cm,width=7cm]{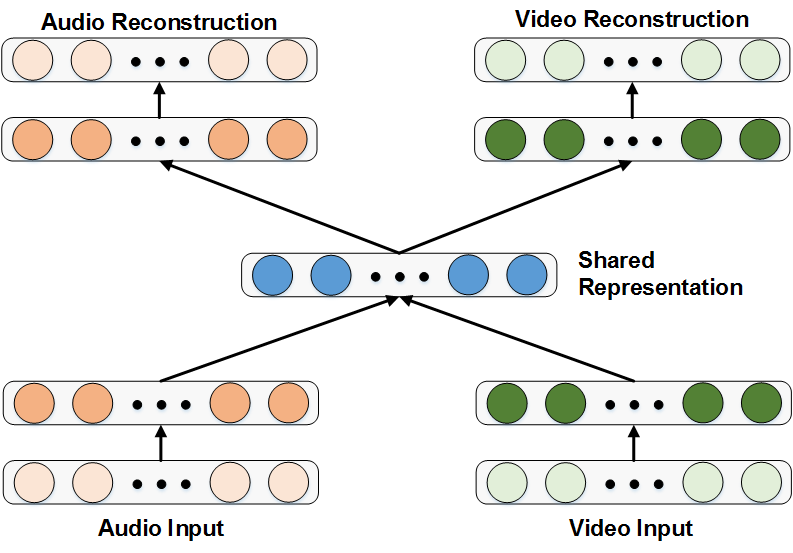}
\end{center}
\caption{The bimodal deep autoencoder (adapted from \cite{ngiam2011multimodal}).}
\label{mDAE}
\end{figure}

To learn features over multiple modalities, Ngiam et al. \cite{ngiam2011multimodal} propose to extract shared representations via training a bimodal deep autoencoder (Figure \ref{mDAE}), which exploits the concatenated final hidden codings of audio and video modalities as input and maps these inputs to a shared representation layer. This fused representation allows the autoencoder to model the relationship between the two modalities. Inspired by denoising autoencoders \cite{vincent2008extracting}, the bimodal autoencoder is trained using an augmented but noisy dataset. Given two-view audio and video dataset $X$ and $Y$, the loss function on any pair of inputs is then defined as follows:
\begin{align}
L(x_i,y_i;~\theta)=L_{I}(x_i,y_i;~\theta)+L_{T}(x_i,y_i;~\theta)
\label{corrAE}
\end{align}
where $L_{I}$ and $L_{T}$ are the losses caused by data reconstruction errors for the given inputs $\{X,Y\}$, specifically audio and video modality. A commonly considered reconstruction loss is the squared error loss,
\begin{align*}
&L_{I}(x_i,y_i;~\theta)=\|x_i-\hat{x}_i\|_2^2\\ 
&L_{T}(x_i,y_i;~\theta)=\|y_i-\hat{y}_i\|_2^2
\end{align*}
As shown in Figure \ref{mDAE}, after having a suitable training stage with bimodal inputs, this model has the ability of using the data from one modality to recover the missing data from the other at test stage. Cadena et al. \cite{cadena2016multi} apply this insight to fuse information available from cameras and depth sensors and reconstruct other missing data for scene understanding problems. 

Consequently, Silberer and Lapata \cite{silberer2014learning} train stacked multi-modal autoencoder with semi-supervised objective to learn grounded meaning representations. In particular, they propose to add a softmax output layer on top of the bimodal shared representation layer to incorporate the object label information. This additional supervised setup is capable of learning more discriminating representations, allowing the network to adapt to specific tasks such as object classification.

Recently, Rastegar et al. \cite{rastegar2016mdl} suggest to exploit the cross weights between representations of modalities for gradually learning interactions of the modalities in a multi-modal deep autoencoder network. Theoretical analysis shows that considering these interactions in deep network manner (from low to high level) provides more intra-modality information. As opposed to the existing deep multi-modal autoencoders, this approach attempts to reconstruct the representation of each modality at a given level, with the representation of the other modalities in the previous layer.

\subsubsection{Multi-View Convolutional Neural Network}
Convolutional neural networks (CNNs) have achieved great success in visual recognition \cite{krizhevsky2012} and this prosperity has been transferred to speech recognition \cite{abdel2014convolutional} and natural language processing \cite{kim2014convolutional} in recent years. In contrast to single-view CNN, multi-view CNN considers learning convolutional representations (features) in the setting in which multiple views of data are available, such as $3$D object recognition \cite{su2015multi}, video action recognition \cite{feichtenhofer2016convolutional}, and person re-identification across multi-cameras \cite{ahmed2015improved}. It seeks to combine the useful information from different views so that more comprehensive representations may be learned for subsequent predictor learning. 

\begin{figure}
\begin{center}
   \includegraphics[scale=0.22]{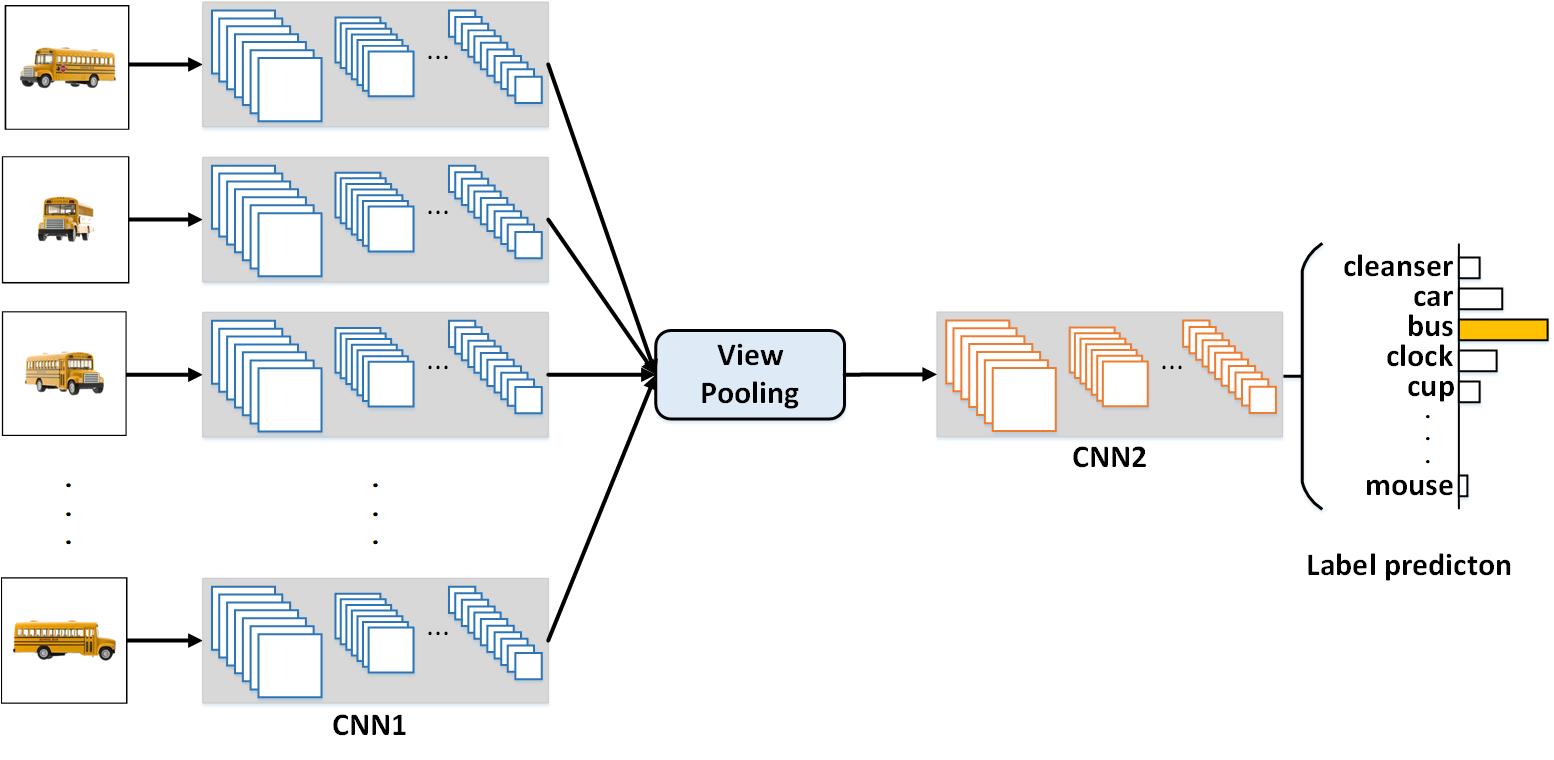}
\end{center}
\vspace*{-0.8em}
\caption{The multi-view CNN architecture (adapted from \cite{su2015multi}).}
\label{mCNN}
\vspace*{-0.8em}
\end{figure}

Taking $3$D object recognition for example, Su et al. \cite{su2015multi} introduce a multi-view CNN architecture that integrates information from multiple 2D views of an object into a single and compact representation. As shown in Figure \ref{mCNN}, multi-view images of a bus with 3D rotations (provided by \cite{kanezaki2018_rotationnet}) are passed through a shared CNN (CNN1) separately, fused at a view-pooling layer, and then sent through the subsequent part of the network (CNN2). Element-wise maximum operation is performed across the views in the view-pooling layer. This multi-view mechanism acts like "data augmentation" where transformed copies of data are added during training to learn the invariance to the alterations such as flips, translations, and rotations. Different from the traditional averaging of final scores from multiple views, this multi-view CNN learns a fused multi-view representation for 3D object recognition.

Further, Feichtenhofer et al. \cite{feichtenhofer2016convolutional} investigate various ways of fusing CNN representations both spatially and temporally to fully exploit the informative spatio-temporal information for human action recognition in videos. It establishes that fusing a spatial and temporal network at a convolutional layer is better than fusing at softmax layer, causing a substantial saving in parameters without loss of performance. Let us take the spatial fusion for example to show its superiority of capturing spatial correspondence. Suppose that $\mathbf{x}^{a}\in \mathbb{R}^{H\times W\times D}$ and $\mathbf{x}^{b}\in \mathbb{R}^{H\times W\times D}$ are two learned feature maps by CNN from two different views $a$ and $b$, respectively.
The proposed {\em conv fusion} first concatenates the two feature maps at the same spatial locations $i$, $j$ across the feature channels $d$ as
\begin{align}
\mathbf{y}^{\text{cat}}=f^{\text{cat}}(\mathbf{x}^{a},\mathbf{x}^{b})
\end{align}
where $y^{\text{cat}}_{i,j,2d}=x^{a}_{i,j,d}$ and $y^{cat}_{i,j,2d-1}=x_{i,j,d}^{b}$. And then the concatenated representation is subsequently convolved with a bank of filters $\mathbf{f}$ and biases $b$,
\begin{align}
\mathbf{y}^{conv}=\mathbf{y}^{\text{cat}}\ast\mathbf{f}+b
\end{align}
where $\mathbf{f}$ is set as $1$D convolution kernel and is responsible for modeling the weighted combinations of the two feature maps $\mathbf{x}^{a}$, $\mathbf{x}^{b}$ at the same spatial location. Consequently, the correspondence between the channels of different views are learned to better classify the actions. 

Multi-view CNN also has been widely applied to the task of person re-identification. Given a pair of images from multiple views as input, this task outputs a similarity value indicating where the two input images represent the same person. Ahmed et al. \cite{ahmed2015improved} introduce a multi-view mid-level feature fusion layer which computes cross-input neighborhood differences to capture local relationships between the two input images. Wang et al. \cite{wang2016joint} categorize the multi-view convolutional representation fusion as a cross-image representation learning and propose a joint learning framework to integrate single-image representation and cross-image representation for person re-identification.
    
\subsubsection{Multi-Modal Recurrent Neural Network}
A recurrent neural network (RNN) \cite{mikolov2010recurrent} is a neural network which processes a variable-length sequence ${\bold x}=(x_{1},\ldots,x_{T})$ through hidden state representation ${\bold h}$. At each time step $t$, the hidden state $h_{t}$ of the RNN is estimated by
\begin{align}
h_{t}=f\left(h_{t-1},x_{t}\right)
\label{rnn}
\end{align}
where $f$ is a non-linear activation function and is selected based on the requirement of data modeling. For example, a simple case may be a common element-wise logistic sigmoid function and a complex case may be a long short-term memory (LSTM) unit \cite{Hochreiter1997}.

An RNN is well-known as its ability of learning a probability distribution over a sequence by being trained to predict the next symbol in a sequence. In this training, the prediction at each time step $t$ is decided by the conditional distribution $p(x_{t}|x_{t-1},\ldots,x_{1})$. For example, a multinomial distribution can be learned as output with a softmax activation function
\begin{align}
p(x_{t,j}=1|x_{t-1},\ldots,x_{1})=\frac{\text{exp}(w_{j}h_{t})}{\sum_{j^{'}=1}^{K}\text{exp}(w_{j^{'}}h_{t})}
\end{align}
where $j=1,\ldots,K$ denotes the possible symbol components and $w_{j}$ are the corresponding rows of a weight matrix $W$. Further, based on the above probabilities, the probability of the sequence ${\bold x}$ can be computed as
\begin{align}
p({\bold x})=\prod_{t=1}^{T}p(x_{t}|x_{t-1},\ldots,x_{1}).
\end{align}
With this learned distribution, it is straightforward to generate a new sequence by iteratively generating a symbol at each time step.

\begin{figure}
\begin{center}
   \includegraphics[height=3.8cm,width=6.7cm]{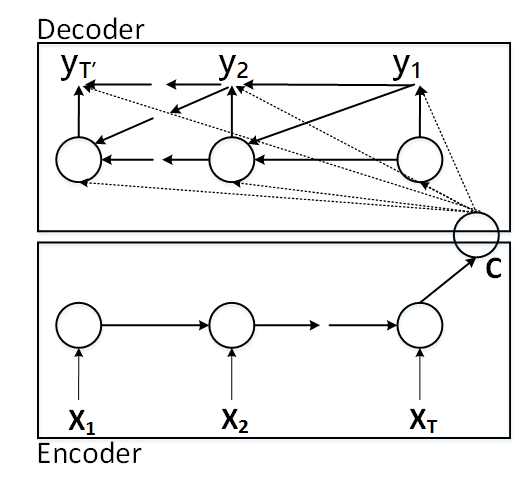}
\end{center}
\caption{The illustration of the RNN encoder-decoder (adapted from \cite{ChoMGBBSB14}).}
\label{RNN}
\end{figure}

Cho et al. \cite{ChoMGBBSB14} propose an RNN encoder-decoder model by exploiting RNN to connect multi-modal sequence. As shown in Figure \ref{RNN}, this neural network first encodes a variable-length source sequence into a fixed-length vector representation and then decodes this fixed-length vector representation back into a variable-length target sequence. In fact, it is a general method to learn the conditional distribution over an output sequence conditioned on another input sequence, e.g., $p(y_{1},\ldots,y_{T^{'}}|x_{1},\ldots,x_{T})$, where the input and output sequence lengths $T$ and $T^{'}$ can be different. In particular, the encoder of the proposed model is an RNN which sequentially encodes each symbol of an input sequence ${\bold x}$ into the corresponding hidden state according to Eq.(\ref{rnn}). After reading the end of the input sequence, a summary hidden state of the whole source sequence ${\bold c}$ is acquired. The decoder of the proposed model is another RNN which is exploited to generate the target sequence by predicting the next symbol $y_{t}$ with the hidden state $h_{t}$. Based on the recurrent property, both $y_t$ and $h_{t}$ are also conditioned on $y_{t-1}$ and on the summary ${\bf c}$ of the input sequence. Thus, the hidden state of the decoder at time $t$ is computed by,
\begin{align}
h_{t}=f(h_{t-1},y_{t-1},{\bold c})
\end{align}              
and the conditional distribution of the next symbol is 
\begin{align}
p(y_{t}|y_{t-1},y_{t-2},\ldots,y_{1},{\bold c})=g(h_{t},y_{t-1},{\bold c})
\end{align}
where $g$ is an activation function and produces valid probabilities with a softmax.
The main idea of the RNN-based encoder-decoder framework can be summarized by jointly training two RNNs to maximize the conditional log-likelihood
\begin{align}
\max_{\theta} \frac{1}{N}\sum_{n=1}^{N}\text{log}p_{\theta}({\bold y_{n}}|{\bold x_{n}})
\end{align}
where $\theta$ is the set of the model parameters and each pair $({\bold x_{n}},{\bold y_{n}})$ consists of an input sequence and an output sequence from the training set. The model parameters can be estimated by a gradient-based algorithm.

Further, Sutskever et al. \cite{SutskeverVL14} also present a general end-to-end approach for multi-modal sequence to sequence learning based on deep LSTM networks, which are very useful for learning problems with long range temporal dependencies \cite{Bengio-trnn94,Hochreiter1997}. The goal of this method is also to estimate the conditional probability $p(y_{1},\ldots,y_{T^{'}}|x_{1},\ldots,x_{T})$. Similar to \cite{ChoMGBBSB14}, the conditional probability is computed by first obtaining the fixed dimensional representation $v$ of the input sequence $(x_{1},\ldots,x_{T})$ with the encoding LSTM-based networks, and then computing the probability of $y_{1},\ldots,y_{T^{'}}$ with the decoding LSTM-based networks whose initial hidden state is set to the representation $v$ of $x_{1},\ldots,x_{T}$:
\begin{align}
p(y_{1},\ldots,y_{T^{'}}|x_{1},\ldots,x_{T})=\prod_{t=1}^{T^{'}}p(y_{t}|v,y_{1},\ldots,y_{t-1})
\end{align}  
where each $p(y_{t}|v,y_{1},\ldots,y_{t-1})$ distribution is represented with a softmax over all the words in the vocabulary.

Besides, multi-modal RNNs have been widely applied in image captioning \cite{KarpathyF14,mao2014deep,kiros2014unifying}, video captioning \cite{venugopalan2014translating,donahue2015long,venugopalan2015sequence}, visual question answering \cite{antol2015vqa}, and information retrieval \cite{PalangiDSGHCSW16}. Karpathy and Li \cite{KarpathyF14} propose a multi-modal recurrent neural network architecture to generate new descriptions of image regions. Chen and Zitnick \cite{chen2014learning} explore the bi-directional mapping between images and their sentence-based descriptions with RNNs. Venugopalan et al. \cite{venugopalan2015sequence} introduce an end-to-end sequence model to generate captions for videos.

By applying attention mechanism \cite{BahdanauCB14} to visual recognition \cite{BaMK14,MnihHGK14}, Xu et al. \cite{XuBKCCSZB15} introduce an attention based multi-modal RNN model, which trains the multi-modal RNN in a deterministic manner using the standard back-propagation. In particular, it incorporates a form of attention with two variants: a "hard" attention mechanism and a "soft" attention mechanism. The advantage of the proposed model lies in attending to salient part of an image while generating its caption.   

\section{Applications}
In general, through exploiting the complementarity of multiple views, multi-view representation learning is capable of learning more informative and compact representation which leads to an improvement in predictors' performance. Thus, multi-view representation learning has been widely applied in numerous real-world applications including cross-media retrieval, natural language processing, video analysis, and recommender system.

\subsection{Cross-media Retrieval} 
As a fundamental statistical tool to explore the relationship between two multidimensional variables, CCA and its extensions have been widely used in cross-media retrieval \cite{Hardoon2004,Socher010,HwangG12}. Hardoon et al. \cite{Hardoon2004} first apply KCCA to cross-modality retrieval task, in which images are retrieved by a given multiple text query without using any label information around the retrieved images. Consequently, KCCA is exploited by Socher and Li \cite{Socher010} to learn a mapping between textual words and visual words so that both modalities are connected by a shared, low dimensional feature space. Further, Hodosh et al. \cite{HodoshYH13} make use of KCCA in a stringent task of associating images with natural language sentences that describe what is depicted.

Inspired by the success of deep learning, deep multi-view representation learning has attracted much attention in cross-media retrieval due to its ability of learning much more expressive cross-view representation. Yu et al. \cite{yu2015learning} present a unified deep neural network model for cross space mapping, in which the image and query are mapped to a common vector space via a convolution part and a query-embedding part, respectively. Jiang et al. \cite{jiang2015deep} also introduce a deep cross-modal retrieval method, called deep compositional cross-modal learning to rank, which considers learning a multi-modal embedding from the perspective of optimizing a pairwise ranking problem while enhancing both local alignment and global alignment. In addition, Wei et al. \cite{wei2016cross} introduce a deep semantic matching method, in which two independent deep networks are learned to map image and text into a common semantic space with a high level abstraction. Wu et al. \cite{wu2016learning} consider learning multi-modal representation from the perspective of encoding the explicit/implicit relevance relationship between the vertices in the click graph, in which vertices are images/text queries and edges indicate the clicks between an image and a query.

\subsection{Natural Language Processing}
There are many Natural Language Processing (NLP) applications of multi-view representation learning. Multi-modal semantic representation models have shown superiority to uni-modal linguistic models on many tasks, including semantic relatedness and predicting compositionally \cite{bruni2014multimodal,CollellZM17,kiela2014learning,roller2013multimodal}. Kiela and Bottou \cite{KielaB14} learn multi-modal concept representations through fusing a skip-gram linguistic representation vector with a deep visual concept representation, which has shown its advantage on tasks of semantic relatedness. Lazaridou et al. \cite{lazaridou2015combining} introduce multimodal skip-gram models to extend the skip-gram model of \cite{mikolov2013efficient} by taking visual information into account. In this extension, for a subset of the target words, relevant visual evidence from natural images is presented together with the corpus contexts.

Further, the idea of multi-view representation fusion has been widely applied in neural network-based sequence-to-sequence (Seq2Seq) learning \cite{SutskeverVL14}. Seq2Seq learning has achieved the state-of-the-art in various NLP tasks, including machine translation \cite{ChoMGBBSB14,BahdanauCB14}, text summarization \cite{rush2015neural,gu2016incorporating}, and dialog system \cite{vinyals2015neural,serban2016building}. It is essentially based on encoder-decoder model where different time steps of each input sequence are fused and encoded into a sequential representation and then a decoder outputs a corresponding sequence from the encoded vector. Taking machine translation for example, Bahdanau et al. encode an input sentence into a sequence of vectors and exploit the fusion of an attentive subset of these vectors to adaptively decode the translation.

\subsection{Video Analysis}
Multi-view representation learning has been used in a number of video analysis tasks, including but not limited to action recognition \cite{yue2015beyond,feichtenhofer2016convolutional}, temporal action detection \cite{yeung2016end,yuan2016temporal}, and video captioning \cite{venugopalan2015sequence,graves2014towards}. For these tasks, representation of different time steps are usually fused to learn a sequential representation which are fed to subsequent predictors. Taking action recognition for example, Ng et al. \cite{yue2015beyond} explore several convolutional temporal feature pooling architectures for video classification. In particular, they perform image feature fusion across a video sequence by employing a recurrent neural network that connects LSTM cells with the outputs of the underlying CNN. Feichtenhofer et al. \cite{feichtenhofer2016convolutional} investigate various ways of fusing CNN representations both spatially and temporally to fully take advantage of the spatio-temporal information. Tran et al. \cite{tran2015learning} propose to learn spatio-temporal features using deep 3D CNN models and show that the learned multi-view fused features encapsulate information related to objects, scenes and actions in a video. 

Following the success of encoder-decoder learning on speech recognition \cite{graves2014towards} and machine translation \cite{SutskeverVL14}, venugopalan et al. \cite{venugopalan2015sequence} propose an end-to-end sequence-to-sequence model to generate descriptions of events in videos. Donahue et al. \cite{donahue2015long} combine convolutional layers and long-range temporal recursion to propose long-term recurrent convolutional networks for visual recognition and description. Ramanishka et al. \cite{ramanishka2016multimodal} introduce a multi-modal video description framework by supplementing the visual information with audio and textual features. This fusion of multiple sources of information shows improvement to exploiting the different modalities separately.

\subsection{Recommender System}
In recommender system, except for the user-item rating information, multiple auxiliary information such as item and user content information can usually be obtained. It is natural to use multi-view representation learning to encode the multiple different sources so that the generalization performance can be improved. Wang and Blei \cite{wang2011collaborative} propose a collaborative topic regression (CTR) model, which seamlessly integrates topic modeling with probabilistic matrix factorization for scientific article recommendation. In particular, CTR produces remarkable and interpretable results through multi-source joint representation learning. 
Consequently, Purushotham et al. \cite{purushotham2012collaborative} propose a hierarchical Bayesian model to connect social network information and item information through shared user latent representation.

With the development of deep learning, collaborative deep learning is first presented by \cite{wang2015collaborative} which integrates stacked denoising autoencoder (SDAE) with probabilistic matrix factorization. It couples SDAE with probabilistic matrix factorization by joint representation learning between rating matrix and auxiliary information. Elkahky et al. \cite{ElkahkySH15} present a multi-view deep representation learning approach for cross-domain user modeling. Dong et al. \cite{dong2017hybrid} propose to jointly perform deep user's and item's latent representation learning from side information and collaborative filtering from the rating data. 

\section{Conclusion}
Multi-view representation learning has attracted much attention in machine learning and data mining areas. This paper introduces two major categories for multi-view representation learning: multi-view representation alignment and multi-view representation fusion. Consequently, we first review the representative methods and theories of multi-view representation learning based on the alignment perspective. Then from the perspective of fusion we investigate the advances of multi-view representation learning that ranges from generative methods including multi-modal topic learning, multi-view sparse coding, and multi-view latent space Markov networks, to neural network models including multi-modal autoencoders, multi-view CNN, and multi-modal RNN. Further, we also discuss several important applications of multi-view representation learning. This survey aims to provide an insightful picture of the theoretical foundation and the current development in the field of multi-view representation learning and to help find the most appropriate methodologies for particular applications.

\section*{Acknowledgments}
This work was supported by National Natural Science Foundation of China (No. 61702448, 61672456), the Fundamental Research Funds for the Central Universities (No. 2017QNA5008, 2017FZA5007), and Zhejiang University --- HIKVision Joint lab. We thank all reviewers for their valuable comments.


%

%

\ifCLASSOPTIONcompsoc
\ifCLASSOPTIONcaptionsoff
  \newpage
\fi



{\small\bibliographystyle{IEEEtran}
\bibliography{IEEEabrv,my}}
%
%
%

%
\begin{IEEEbiography}{Yingming Li} received the BS and MS degrees in automation from University of Science and Technology of China, Hefei, China, and the PhD degree in information science and electronic engineering from Zhejiang University, Hangzhou, China. He is currently an assistant professor with the College of Information Science and Electronic Engineering at Zhejiang University, China.
\end{IEEEbiography}

\begin{IEEEbiography}{Ming Yang} received the BS, MS, and PhD degrees in information science and electronic engineering from Zhejiang University, Hangzhou, China. He had been a visiting scholar in computer science at the State University of New York (SUNY) at Binghamton.
\end{IEEEbiography}



\begin{IEEEbiography}{Zhongfei (Mark) Zhang} received the BS degree in electronics engineering, the MS degree in information science, both from Zhejiang University, Hangzhou, China, and the PhD degree in computer science from the University of Massachusetts at Amherst. He is a QiuShi Chaired Professor at Zhejiang University, China, and directs the Data Science and Engineering Research Center at the university while he is on leave from State University of New York (SUNY) at Binghamton, USA, where he is a professor at the Computer Science Department and directs the Multimedia Research Laboratory in the Department. He has published more than 200 peer-reviewed academic papers in leading international journals and conferences and several invited papers and book chapters, has authored or co-authored two monographs on multimedia data mining and relational data clustering, respectively.

\end{IEEEbiography}
\end{document}